\documentclass[10pt,twocolumn,letterpaper]{article}

\usepackage{wacv}
\usepackage{times}
\usepackage{epsfig}
\usepackage{graphicx}
\usepackage{amsmath}
\usepackage{amssymb}
\usepackage{multirow}
\usepackage{subcaption}
\usepackage{titling}

\def\wacvPaperID{581} 

\wacvfinalcopy 

\ifwacvfinal
\def\assignedStartPage{9876} 
\fi

\ifwacvfinal
\usepackage[breaklinks=true,bookmarks=false]{hyperref}
\else
\usepackage[pagebackref=true,breaklinks=true,colorlinks,bookmarks=false]{hyperref}
\fi

\ifwacvfinal
\else
\fi

\begin{document}

\title{Comprehensive Analysis of the Object Detection Pipeline on UAVs}

\author{Leon Amadeus Varga\\
Cognitive Systems\\
University of Tuebingen, Germany\\
{\tt\small leon.varga@uni-tuebingen.de}
\and
Sebastian Koch\\
Cognitive Systems\\
University of Tuebingen, Germany\\
{\tt\small se.koch@student.uni-tuebingen.de}
\and
Andreas Zell\\
Cognitive Systems\\
University of Tuebingen, Germany\\
{\tt\small andreas.zell@uni-tuebingen.de}
}

\maketitle

\begin{abstract}
An object detection pipeline comprises a camera that captures the scene and an object detector that processes these images. The quality of the images directly affects the performance of the object detector. 
Many works nowadays focus either on improving the image quality or improving the object detection models independently, but neglect the importance of joint optimization of the two subsystems. The goal of this paper is to tune the detection throughput and accuracy of existing object detectors in the remote sensing scenario by focusing on optimizing the input images tailored to the object detector.
To achieve this, we empirically analyze the influence of two selected camera calibration parameters (camera distortion correction and gamma correction) and five image parameters (quantization, compression, resolution, color model, additional channels) in remote sensing applications. 
For our experiments, we utilize three UAV data sets from different domains and a mixture of large and small state-of-the-art object detector models to provide an extensive evaluation of the influence of the pipeline parameters. 
Finally, we realize an object detection pipeline prototype on an embedded platform for an UAV and give a best practice recommendation for building object detection pipelines based on our findings. We show that not all parameters have an equal impact on detection accuracy and data throughput, and that by using a suitable compromise between parameters we are able to achieve higher detection accuracy for lightweight object detection models, while keeping the same data throughput.
\end{abstract}


\begin{figure}
    \centering
    \includegraphics[width=0.55\linewidth]{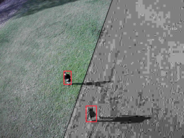}
    
    \vspace{2px}
      
    \includegraphics[width=0.55\linewidth]{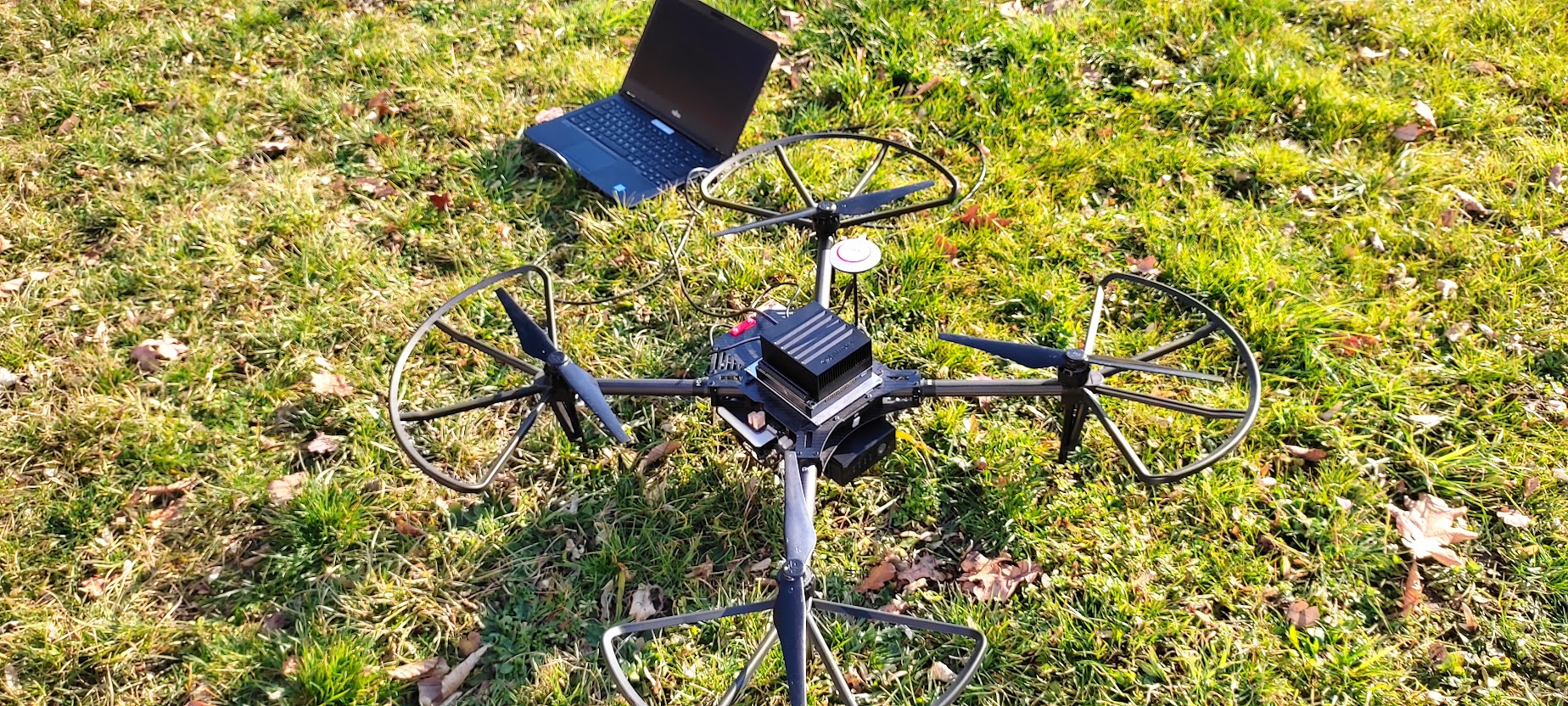}
    \caption{Prototype system using an optimized detection pipeline.}
    \label{fig:drone_system}
    \vspace{-1.5em}
\end{figure}

\section{Introduction}
Micro Aerial Vehicles (MAVs) and Unmanned Aerial Vehicles (UAVs) can support surveillance tasks in many applications. Examples are traffic surveillance in urban environments \cite{DBLP:journals/corr/abs-2001-06303}, or search and rescue missions in maritime environments \cite{Varga2021a}, or the much more general task of remote sensing object detection \cite{Ding2021}. Two key components of these tasks are the camera system and the object detector. The interaction, which is normally based on image data, of these components is crucial. This work analyzes the impact of image data settings. This kind of analysis is currently only available for the advanced driver assistance system (ADAS) scenario (see section \ref{link:related_work}).\\ 
In most projects, the configuration of these components has to be conducted at an early stage. A recording of a necessary data set should be already done with the final camera system. This leads to a chicken-and-egg problem. The optimal configuration depends highly on the application, but we provide some intuitions of the general behavior for the remote sensing use case.\\
The emerging trend towards smart cameras, combining the camera and the computing unit, confirms the need for classification and detection pipelines. The lack of flexibility and the high price of these devices are still a serious drawback and prevent their use in research or prototyping. In this work, the detection pipeline is analyzed without smart cameras, but the results could also be useful for projects that include them. \\ 
Besides the experiments (in a desktop and embedded environment) and the result analysis, we applied the outcomes to build a proof-of-concept system. In this work, we provide the most important findings. Additionally, in our supplementary, we provide further details on the used detection models and datasets, the code and a full list of experiments.
Our contributions are:

\begin{itemize}

    \item We evaluated the impact of \textbf{7 parameters} on three different data sets and \textbf{four models}. 
    \item We achieve \textbf{improved detection accuracy with similar data throughput} on an embedded platform and provide a best practice recommendation for the evaluated parameters for the UAV object detection.
    \item Finally, we demonstrate the usability with a \textbf{prototype}.

\end{itemize}

\begin{figure}
 \begin{subfigure}{0.30 \columnwidth}
        \centering
        \includegraphics[height=3.3cm]{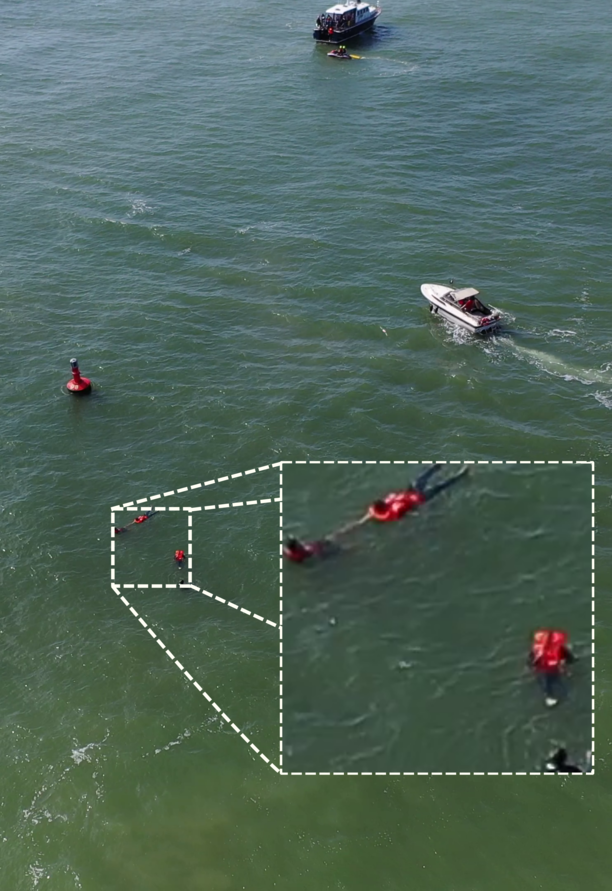}
        \caption{8 Bit}
    \end{subfigure}
    \begin{subfigure}{0.30 \columnwidth}
        \centering
        \includegraphics[height=3.3cm]{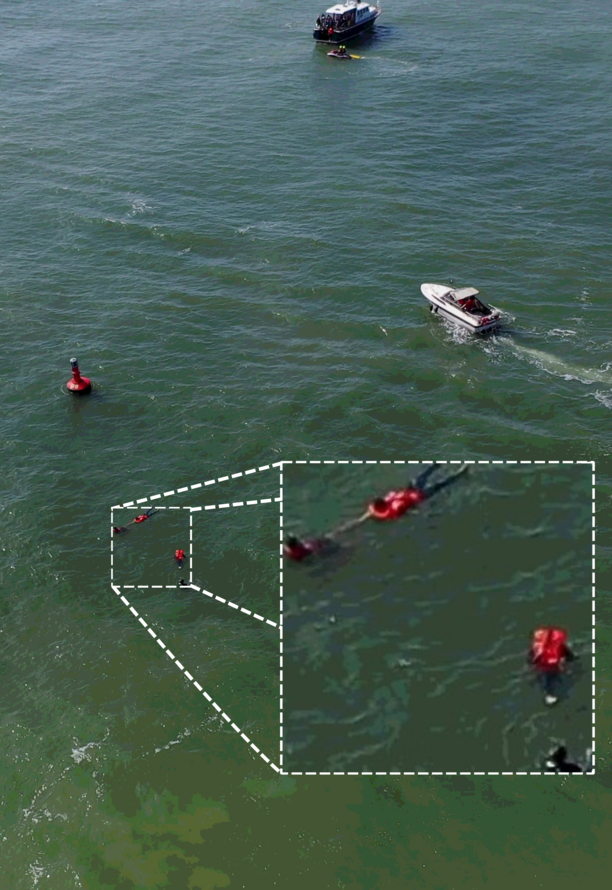}
        \caption{4 Bit}
    \end{subfigure}
    \begin{subfigure}{0.30 \columnwidth}
        \centering
        \includegraphics[height=3.3cm]{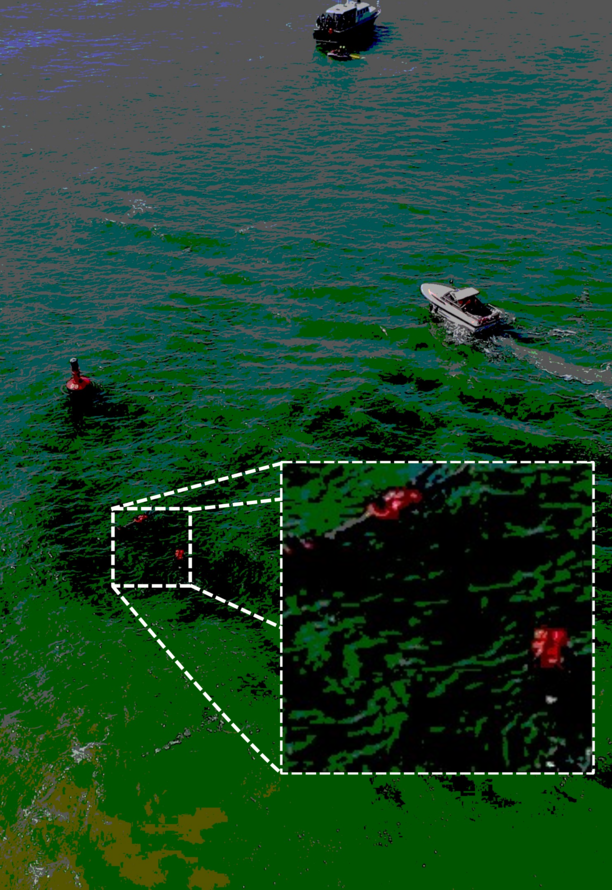}
        \caption{2 Bit}
    \end{subfigure}
    \centering
    \caption{Example images for Quantization}
    \label{fig:example_quantization}
\end{figure}

\section{Related work}
\label{link:related_work}
The image processing pipeline has a large impact on the performance of object detectors. Many parameters influence this pipeline. However, there is still a lack of impact analysis for UAV-based systems. Yahyanejad \etal has measured the impact of lens distortion correction for thermal images acquired by a UAVs \cite{DBLP:conf/rose/YahyanejadMR11}. However, their results are only useful for a thermal camera system. \\
Other existing camera parameter evaluations are mostly based on the autonomous driving scenario \cite{Blasinski2018OptimizingIA, DBLP:conf/eccv/CarlsonSVJ18,  DBLP:conf/iccvw/LiuLFW19, DBLP:journals/access/LiuLFW20, DBLP:conf/iccve/SaadS19, DBLP:conf/issre/SecciC20}. The findings of the autonomous driving experiments cannot be directly applied to the remote sensing use-case. For autonomous driving, the scene is often homogenous, and the object sizes differ only slightly. Further, the distance to the objects is often only a couple of meters. For remote sensing recordings, the distance to the scene is often over 30 meters \cite{Varga2021a}, resulting in tiny objects. Depending on the camera angle, the images can vary strongly. Because of these differences, the results of one scenario are not completely valid in the other, but can give a first sign.\\
Balsinski \etal proposed a render framework to evaluate camera architectures in simple autonomous driving scenarios \cite{Blasinski2018OptimizingIA}. Secci \etal evaluated camera failures in the autonomous driving application \cite{DBLP:conf/issre/SecciC20}. Besides these, the works of Liu \etal\cite{DBLP:conf/iccvw/LiuLFW19}, Carlson \etal\cite{DBLP:conf/eccv/CarlsonSVJ18} and Saad \etal\cite{DBLP:conf/iccve/SaadS19} showed the still existing gap between synthetic and real-world data. Thus, our experiments are based entirely on real-world recordings.\\
Buckler \etal analyzed the image pipeline regarding performance and energy consumption \cite{DBLP:conf/iccv/BucklerJS17}. In contrast to their work, we close the gap for a missing camera parameter evaluation in the remote sensing application. Further, we consider parameters, which are configurable for most off-the-shelf camera systems.\\
In contrast to work that accelerates inference by, e.g., quantizing the neural network \cite{DBLP:conf/cvpr/LiWLQYF19, DBLP:conf/cvpr/CaiYDGMK20}, our work focuses on the parameters of the object recognition pipeline that affect the overall throughput time of the system, rather than on the inference time itself.\\
In summary, our work differs from the aforementioned works in the following aspects. We focus on the remote sensing use-case for UAVs and instead of using synthetic data, we used established real-world data sets. The evaluated camera parameters are configurable for most cameras.

\section{Considered factors of influence}
Our goal is to evaluate the impact of adjustable acquisition parameters on the performance of object detectors in UAV remote sensing. We selected parameters which have a direct influence on the recordings. Three parameters, \textit{Quantization}, \textit{Compression} and \textit{Resolution}, control the required memory consumption. Further, we evaluated the impact of thorough camera setup and calibration. We select radial \textit{Image Distortion} which originates from the camera lens and \textit{Gamma Correction} to simulate different exposure settings for a prerecorded dataset. Other camera effects and calibration parameters like focus distance are also interesting, but harder to simulate with prerecorded dataset. Thus, we limit ourselves to those chosen effects and measure the impact of a correctly exposed and distortion corrected camera. Further, we evaluate the impact of the selected \textit{Color Model}. \textit{Multispectral recordings} can carry useful additional information; we therefore check how easily they can be integrated into existing architectures and whether they help to improve the prediction capability.\\
We limit our evaluation to these parameters because we believe that these parameters have a direct influence on the recording and can be simulated on existing datasets. 

\subsection{Quantization}
The quantization, also called bit depth for images, defines the number of different values per pixel and color (see Fig. \ref{fig:example_quantization}). Quantization is used to describe a continuous signal with discrete values \cite{1089500}. Too few quantization steps lead to a large quantization error, which results in a loss of information. Therefore, it is necessary to find a balance between the required storage space and the information it contains.\\
All used data sets provide images with a bit depth of 8bit per color channel. We want to evaluate, whether we can reduce the bit depth further without losing performance. This could reduce the data size. Hence, 8bit is the maximal available quantization of the data sets. We provide experiments for 8bit, 4bit and 2bit quantization, which would decrease the size of the images for a bitmap format by factor 2 (for 4bit) and 4 (for 2bit).

\begin{figure}
 \begin{subfigure}{0.30 \columnwidth}
        \centering
        \includegraphics[height=3.3cm]{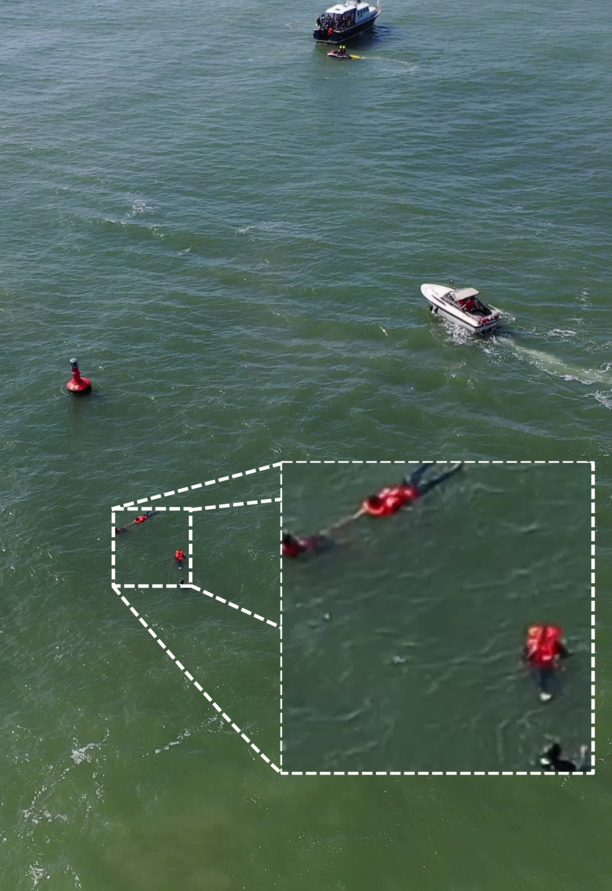}
        \caption{No comp.}
    \end{subfigure}
    \begin{subfigure}{0.30 \columnwidth}
        \centering
        \includegraphics[height=3.3cm]{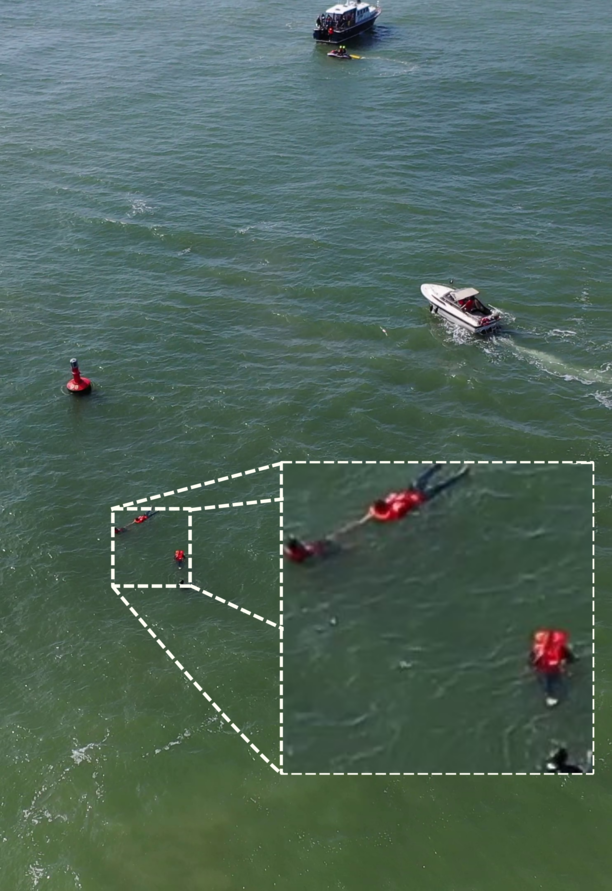}
        \caption{90 comp. qlty.}
    \end{subfigure}
    \begin{subfigure}{0.30 \columnwidth}
        \centering
        \includegraphics[height=3.3cm]{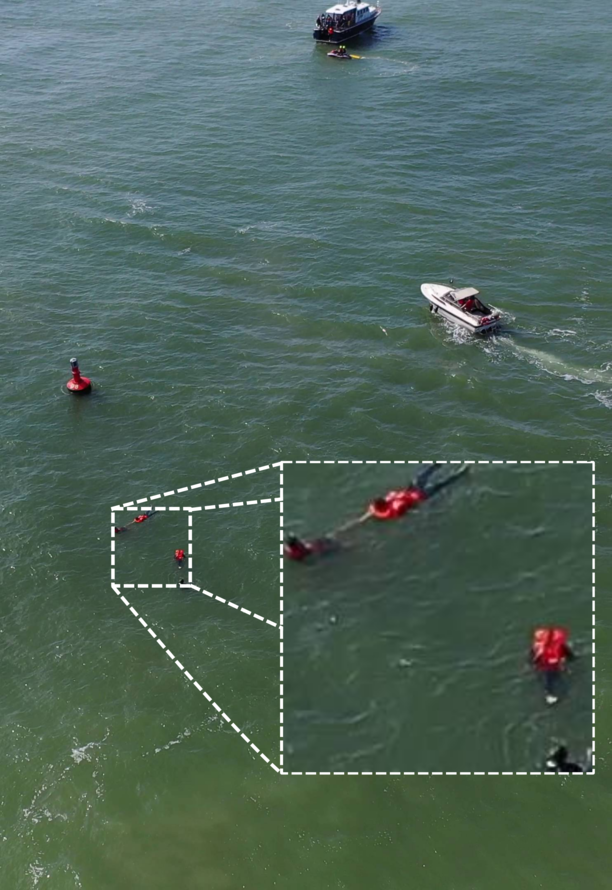}
        \caption{70 comp. qlty.}
    \end{subfigure}
    \centering
    \caption{Example images for Compression}
    \label{fig:example_compression}
\end{figure}

\subsection{Compression}
Compression is used to reduce the required space of a recording. There are two types of compressions: lossless and lossy compressions. The image format PNG belongs to lossless compressions techniques. The compression time for PNG is very high, therefore it is often not considered for online compression. JPEG is faster, but also a lossy compression. The compression removes information. Nevertheless, JPEG is a common choice. It provides a sufficient trade-off between loss of information, file size and speed. \\
The compression quality of JPEG is defined as a value between 1 and 100 (default: 90). Higher values result in less compression.\\
We check the relation between compression and detector performance. A lower compression quality would cause smaller file sizes. This could directly lead to a higher throughput of the system. For the inferences of models, the decompressed data is used. A lower compression quality would, therefore, mainly affect the throughput from the camera to the processing unit memory. Since this is also a limiting factor for high-resolution or high-speed cameras, it is worth checking. In our experiments, we evaluate the influence of no compression and a compression quality for JPEG of 90 and 70  (see Fig. \ref{fig:example_compression}).

\begin{figure}
 \begin{subfigure}{0.30 \columnwidth}
        \centering
        \includegraphics[height=3.3cm]{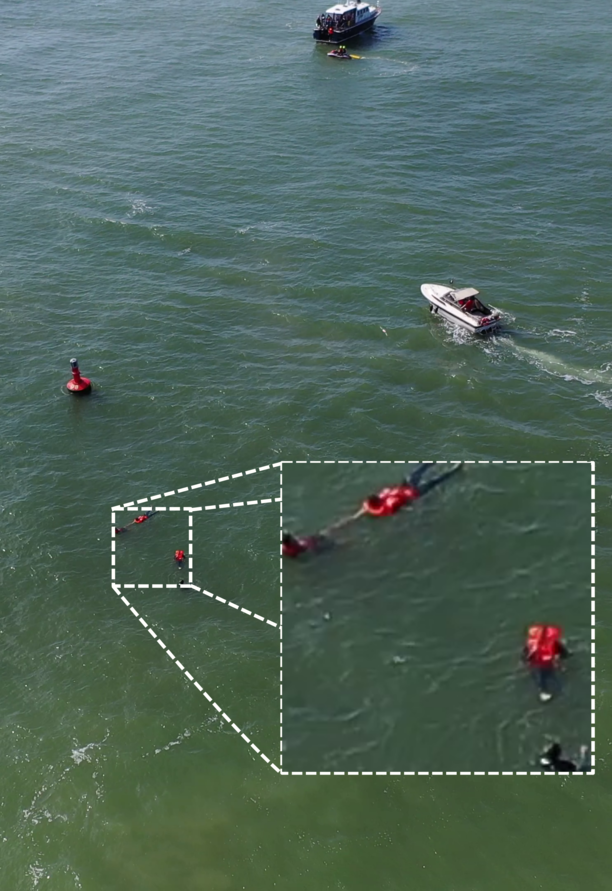}
        \caption{Full resolution}
    \end{subfigure}
    \begin{subfigure}{0.30 \columnwidth}
        \centering
        \includegraphics[height=3.3cm]{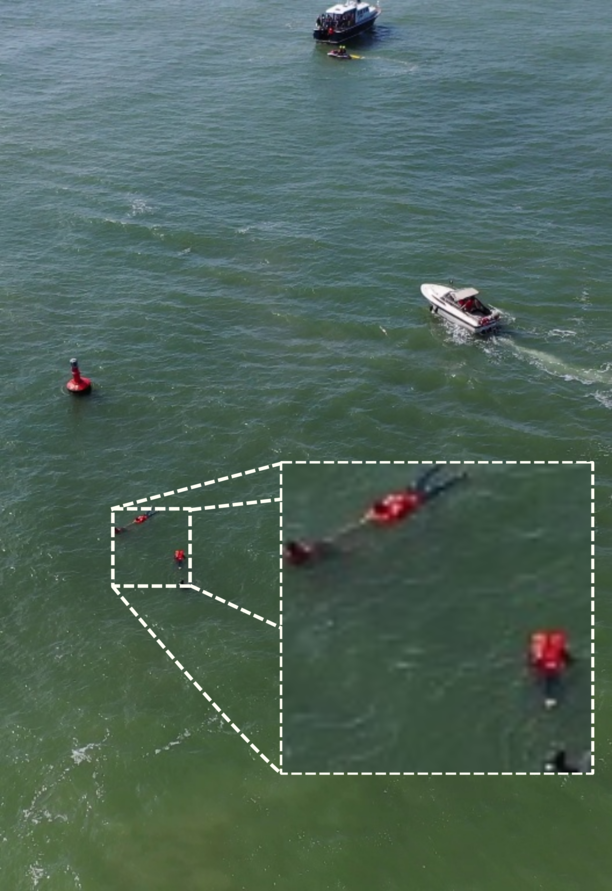}
        \caption{Max 768 pixels}
    \end{subfigure}
    \begin{subfigure}{0.30 \columnwidth}
        \centering
        \includegraphics[height=3.3cm]{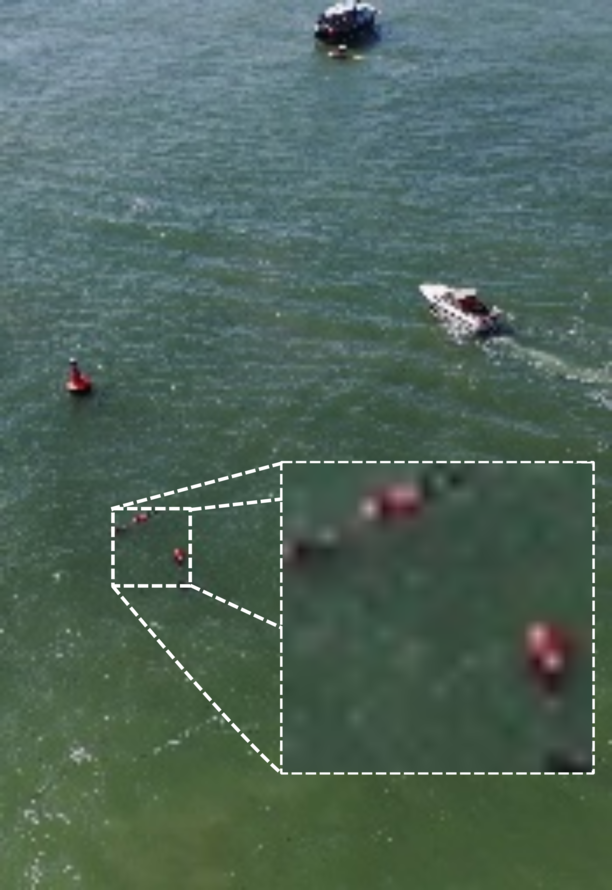}
        \caption{Max 256 pixels}
    \end{subfigure}
    \centering
    \caption{Example images for image resolutions with max side length}
    \label{fig:example_resolution}
\end{figure}

\subsection{Color Model}
The color model defines the representation of the colors. One of the most common one is RGB \cite{hunt2005reproduction}. It is an additive color model, which is based on the three colors red, green, and blue. For object detection and object classification, this representation is mostly used \cite{DBLP:journals/pami/RenHG017, DBLP:conf/cvpr/TanPL20}. Besides RGB, non-linear transformations of RGB are also worth considering. HLS and HSV are based on a combination of hue and saturation. Both differ only in the brightness's definition value. YCbCr takes human perception into account and encodes information more suitably for the human eye.\\
In some applications, HSV outperforms RGB. Cucchiara \etal \cite{948679} and Shuhua \etal \cite{5497299} could confirm this performance improvement. Kim \etal demonstrated RGB outperforms the other color models (HSV, YCbCr, and CIE Lab) in traffic signal detection. It appears, which color model is best depends strongly on the application and the model.\\
The color model is also a design decision, so we evaluate the performance of the RGB, HSV, HLS, and YCbCr color models for object detection in remote sensing. We also review the performance of gray-scale images. To achieve comparable results, the backbones are not pre-trained for these experiments.

\subsection{Resolution}
Higher resolutions reveal more details, but also result in larger recordings. Especially for the small objects of remote sensing, the details are crucial \cite{DBLP:journals/lgrs/LiuM15}. Therefore, choosing an optimal resolution is important \cite{DBLP:conf/icml/TanL19}.\\
Besides the maximal available resolution defined by the used camera, the GPU memory often limits the training size. The reason lies in the gradient calculation within the backpropagation step. There are techniques, which can reduce this problem \cite{Varga_2021_ICCV, messmer2021gaining}. To allow a fair comparison, we limit the maximum training size to 1024 pixels for the length of the larger side. This training size is processable with all models used. The experiments should show the performance reduction for smaller image sizes. So we trained on the training sizes of 256 pixels, 512 pixels, 756 pixels and 1024 pixels (see examples in Fig. \ref{fig:example_resolution}). Further, we evaluated each training size with three test sizes, which are defined by the training size $\times$ the factors 0.5, 1 and 2.

\subsection{Multispectral Recordings}
\begin{figure}
 \begin{subfigure}{0.47 \columnwidth}
        \centering
        \includegraphics[height=3.2cm]{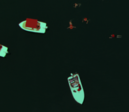}
        \caption{RGB}
    \end{subfigure}
    \begin{subfigure}{0.47 \columnwidth}
        \centering
        \includegraphics[height=3.2cm]{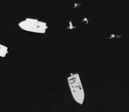}
        \caption{Near-infrared}
    \end{subfigure}
    \centering
    \caption{Example images for multispectral recordings in maritime environment. In the NIR recordings, the swimmers are fully visible.}
    \label{fig:example_multispectal}
    \vspace{-1em}
\end{figure}

In recent years, multispectral cameras got more popular for usage on drones \cite{candiago2015evaluating} \cite{deng2018uav} \cite{zhang2019mapping}. The additional channels of these increase the recording size, but can also provide important information. In contrast to conventional RGB cameras, multispectral cameras add channels outside the visible light (e.g. near-infrared or thermal). In pedestrian and traffic monitoring, multispectral imaging can increase the detector performance, as shown by Takumi \etal \cite{DBLP:conf/mm/TakumiWHTUH17}, Maarten \etal \cite{DBLP:conf/iciar/VandersteegenBG18}, and Dayan \etal\cite{DBLP:journals/inffus/GuanCYCY19}. And for precision agriculture, they are one of the key components \cite{candiago2015evaluating} \cite{deng2018uav} \cite{zhang2019mapping}. We want to create awareness, that multispectral recordings can be helpful in some situation. But we also want to check how easy it is to integrate these into existing object detection pipelines. So, the focus is not on evaluating multimodal models \cite{DBLP:journals/sensors/OphoffBG19}, and we consider multispectral images as a limiting case to classical color images. Our experiments are limited to the early fusion approach \cite{DBLP:journals/ivc/ZhangSMM21}.\\
We used the multispectral images from the SeaDronesSee \cite{Varga2021a} data set for our experiments. These were acquired with a MicaSense RedEdge MX and thus provide wavelengths of 475 nm (blue), 560 nm (green), 668 nm (red), 717 nm (red edge), and 842 nm (near infrared). We evaluate how the additional wavelengths affect the performance of the object detectors in detecting swimmers (see example in Fig. \ref{fig:example_multispectal}). The trend of the results should be generally usable, even if the exact results are only for the maritime environment. Always, if additional channels could carry useful information, it is worth considering an evaluation of this kind.

\subsection{Calibration Parameters}
\paragraph{Image Distortion:}
During the image formation process, most lenses introduce a distortion in the digital image that does not exist in the scene \cite{ISO17850:2015}. Wide-angle lenses or fish-eye lenses suffer from the so-called barrel-distortion. 
Telephoto lenses suffer from the so-called pincushion-distortion.\\
Often, one of the first steps in the image processing pipeline is to correct the lens distortion using a reference pattern.\\
As we are working on data sets that are distortion rectified, we artificially introduce distortion via the Brown–Conrady model \cite{brown}. To evaluate the impact of an incorrect calibration, we introduce a small barrel-distortion with $k_1 = - 0.2$ and pincushion-distortion with $k_1 = 0.2$. A visualization of these distortions can be found in the supplementary material.

\paragraph{Gamma Correction:}
For most cameras use an auto-exposure setting which adjusts the camera exposure to the ambient lighting. But manual exposure is still used and can result in overexposed (to bright) or underexposed (to dark) images. Gamma correction can to an extent ease this problem, but it can also be used to simulate a camera in a wrong exposure setting.\\
Gamma Correction is a non-linear operation to encode and decode luminance to optimize the usage of bits when storing an image, by taking advantage of the non-linear manner in which humans perceive light and color \cite{poynton2012digital}. Gamma Correction can also simulate a shorter or longer exposure process after the image formation process. 
If images are not gamma-corrected, they allocate too many bits for areas that humans cannot differentiate. Neural networks, however, differ from humans in the sense that they work with floating-points which makes them sensitive to minimal changes in quantization as can be observed in many works that examine weight quantization \cite{zafrir2019q8bert, krishnamoorthi2018quantizing, jacob2018quantization}. Contrarily, humans perceive changes in brightness logarithmically \cite{SHEN2003241}.  
We therefore assume that Gamma Correction may have a negligible impact on neural network detection performance. \\
Many works have employed Gamma Correction, intending to improve detection performance directly or help training with Gamma Correction as an augmentation technique \cite{wang2019augpod, zhang2019real}.\\
We use three different gamma configurations (0.5, 1.0, 2.5) to investigate the effects of gamma on the object detector performance. These experiments are designed to show whether gamma is a crucial parameter. The performance of a dynamic gamma adjustment \cite{4266947} will also be tested.

\begin{figure*}
    \centering
    \includegraphics[width=0.8\textwidth]{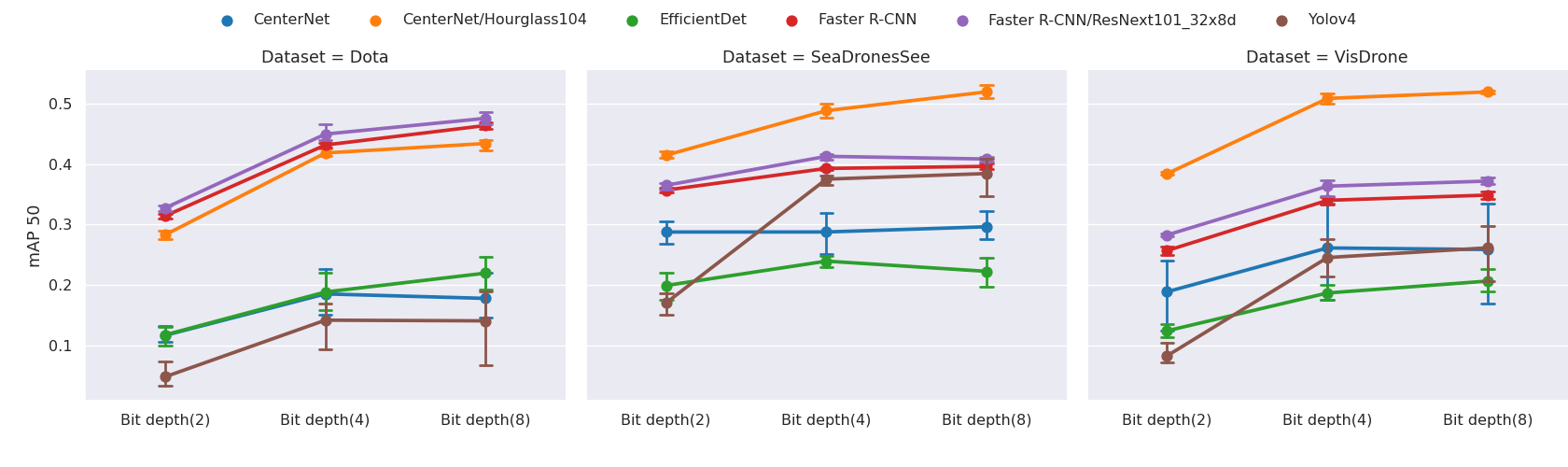}
    \caption{Quantization: the impact of the quantization on the performance of the object detectors for all data sets. The y-axis shows the mean Average Precision (mAP).}
    \label{fig:quanitzation}
\end{figure*}

\section{Experiments}
In this section, we describe the setup of the experiments. The setup contains the data sets, the object detectors and the hardware environment. For each configuration, we trained with three different random seeds. The random seed affects the weight initialization, the order of the training samples and data augmentation techniques. The three initializations show the stability of each configuration.

\subsection{Datasets}
We evaluated our experiments on three data sets. All data sets are part of the remote sensing scenario. Dota-2 is based on satellite recordings \cite{Ding2021}. VisDrone \cite{DBLP:journals/corr/abs-2001-06303} and SeaDronesSee \cite{Varga2021a} were recorded by Micro Air Vehicles (MAVs). The first two data sets are well established and commonly used to benchmark object detection for remote sensing. In contrast, SeaDronesSee focuses on a maritime environment, which differs from the urban environment of VisDrone and Dota-2. Therefore, it introduces new challenges and is a proper extension for our experiments. Further details can be found in the supplementary material.

\begin{figure*}
    \centering
    \includegraphics[width=0.8\textwidth]{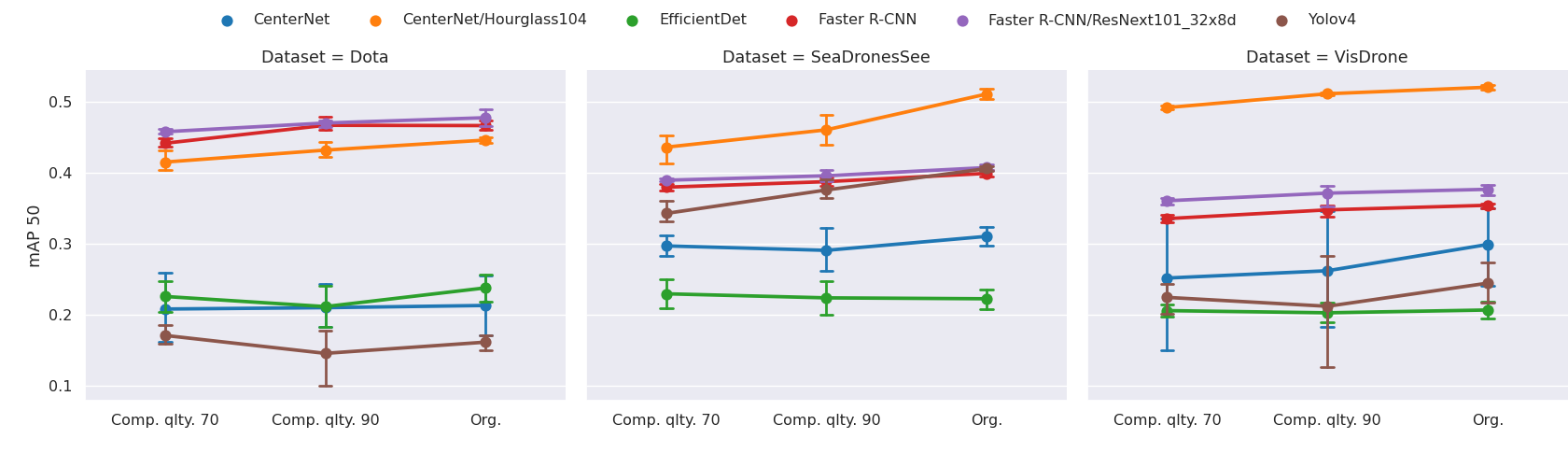}
    \caption{Compression: the impact of the JPEG compression on the performance of the object detectors for all data sets. The y-axis shows the mean Average Precision (mAP). A higher compression quality means less compression.}
    \label{fig:compression}
\end{figure*}

\subsection{Models}
For our experiments, we selected four different object detectors. The two-stage detector Faster-RCNN \cite{DBLP:journals/pami/RenHG017} can still achieve state-of-the-art results with smaller modifications \cite{DBLP:journals/corr/abs-2001-06303}. By comparison, the one-stage detectors are often faster than the two-stage detectors. Yolov4 \cite{Bochkovskiy2020} and EfficientDet \cite{DBLP:conf/cvpr/TanPL20} are one-stage detectors and focus on efficiency. Both perform well on embedded hardware. Modified versions of CenterNet \cite{Duan2019}, also a one-stage detector, could achieve satisfying results in the VisDrone challenge \cite{DBLP:journals/corr/abs-2001-06303}. These four models cover a variety of approaches and can give a reliable impression of the parameter impacts.\\
For Faster R-CNN, EfficientDet and CenterNet, we used multiple backbones to provide an insight for different network sizes. The networks and their training procedure are described in the supplementary material in further detail.

\subsection{Hardware setup}
Most of the experiments were done in a desktop environment. To evaluate the performance in a more applicable environment, we performed experiments on an Nvidia Xavier AGX board. The small size and the low weight allow the usage for onboard processing in MAVs or other robots. The full description of the desktop and embedded environment can be found in the supplementary material.

\section{Results and discussion}
In this section, we present the results of the experiments. Each parameter is considered in its own subsection. The experiments of the different backbones per network were grouped, only the largest backbones of CenterNet and Faster R-CNN are listed separately. As an evaluation metric for the object detectors, we use the mean Average Precision (mAP) with an IoU threshold of 0.5 \cite{DBLP:conf/eccv/LinMBHPRDZ14}.\\
In the Dota dataset, the performance of the models differs from the performance in the other datasets. The objects in the Dota dataset are very small, unlike the other datasets. This requires different properties of the object detectors.
For a more detailed analysis of this behavior and a complete list of all experiments, we refer the reader to supplementary material$^{4}$.
Here we highlight the most important results in terms of parameters.

\subsection{Quantization}
A quantization reduction leads to a loss of information in the recording. This assumption is confirmed by the empirical results in Fig. \ref{fig:quanitzation}. This phenomenon is consistent for all data sets and all models. However, deterioration varies.\\
Worth mentioning is the marginal difference ($\sim 3\%$) in performance between 4bit and 8bit recordings. Hence, it is possible to have half the amount of data and achieve nearly the same result. The loss in performance for a 2bit quantization is much higher ($\sim 41\%$).

\begin{figure*}
    \centering
    \includegraphics[width=0.8\textwidth]{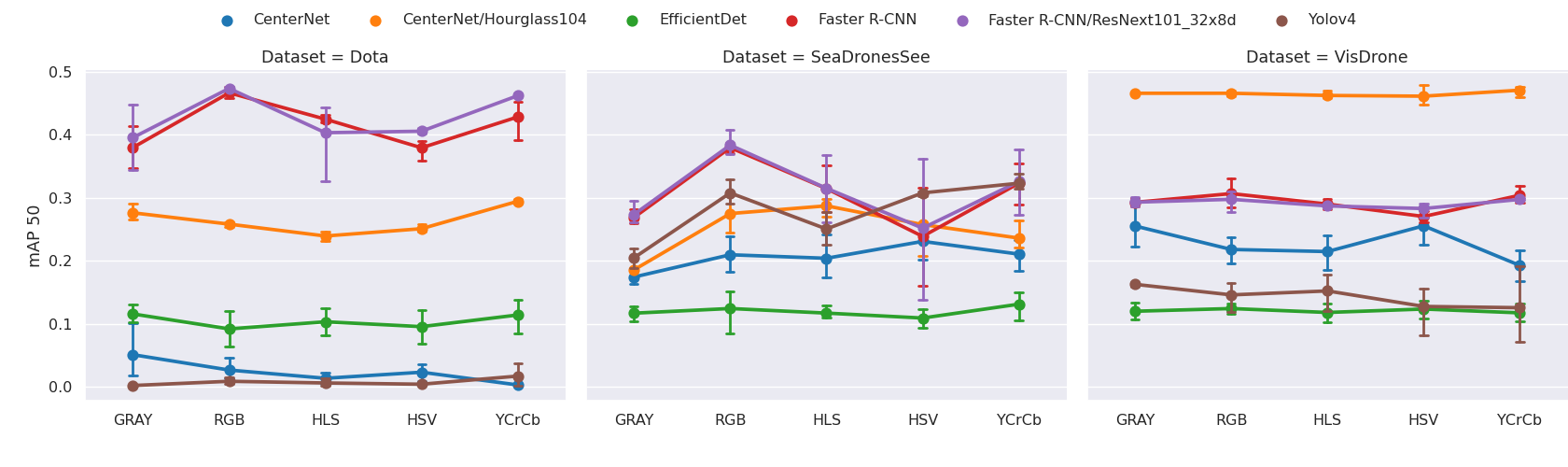}
    \caption{Color Model: the impact of the color models. The y-axis shows the mean Average Precision (mAP).}
    \label{fig:color_space}
\end{figure*}

\begin{figure*}
	\centering
	\includegraphics[width=0.8\textwidth]{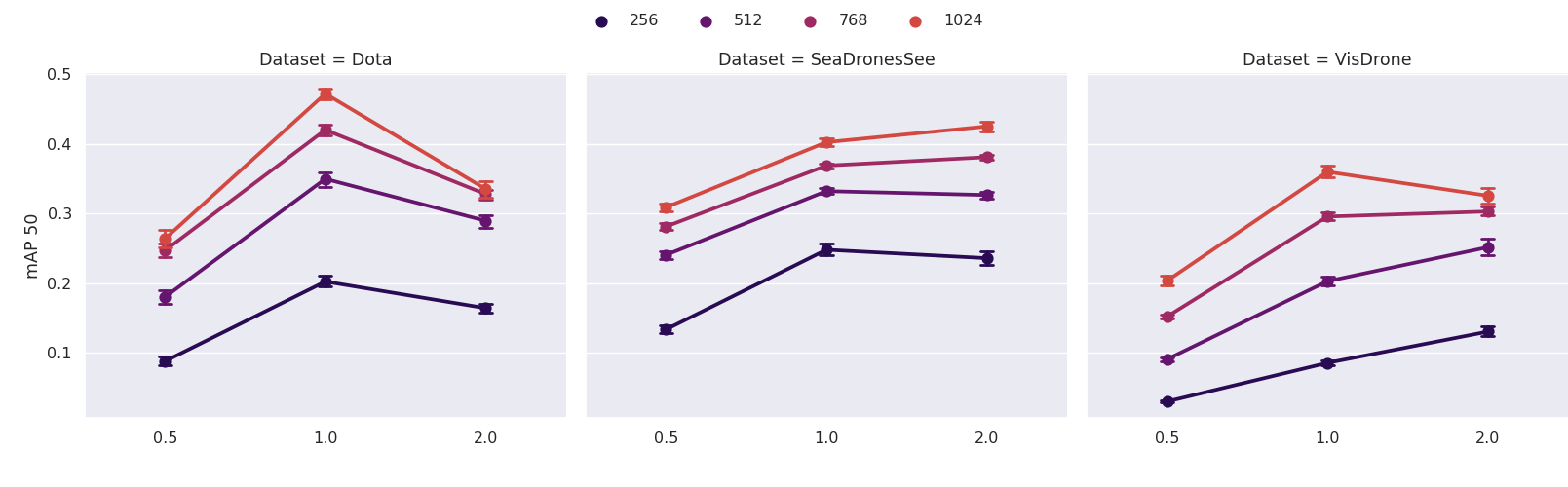}
	\caption{Image Size: Faster R-CNN trained with different training image sizes and validated on $\text{validation image size} = x \times \text{training image size} $.  The y-axis shows the mean Average Precision (mAP).}
	\label{fig:image_size}
\end{figure*}

\subsection{Compression}
Fig. \ref{fig:compression} shows the relation between compression quality and object detector performance. Lower compression quality, i.e. higher compression, leads to worse performance. With some exceptions, this is true for all models. Overall, the performance degeneration of the tested compression qualities is small. For a compression quality of 90, the performance is decreased by $\sim 2.7\%$. A compression quality of 70 leads to an mAP loss of $\sim 5.3\%$. Compared to the performance loss, the size reduction is significant. \\
We observe that a compression quality of 90 reduces the required space by $\sim 25\%$ for a random subset of the VisDrone validation set. A compression quality of 70 can almost halve the required space. \\
However, the memory reduction depends highly on the image content. For SeaDronesSee and Dota the memory reduction is much higher. For compression quality of 70 and 90, it is about $\sim 93\%$ and, $\sim 86\%$ respectively. \\
By reducing the required image space, it is possible to increase the pipeline throughput. In return, a compression and decompression is necessary. In the final experiments (see section \ref{link:application}), the utility of this approach becomes clear.

\subsection{Color Model}
\label{link:color_model}
We can conclude two important outcomes of the experiment for color models. First, we have to distinguish between two situations, which are visible in Fig. \ref{fig:color_space}. In the first situation, the object detectors seem to not take advantage of the color information. Therefore, we assume the color is in this case not beneficial. This applies to the VisDrone data set and for the Dota data set. Both data sets provide objects and backgrounds with a high color variance. For example, a car in the VisDrone data set can have any color. Thus, the texture and shape are much more important. That means the object detector can already achieve excellent results with the gray-scale recordings. For the Dota data set, the color information could only be used by the largest models (Faster R-CNN/ResNext101\_32x8d and CenterNet/Hourglass104). The other models weren’t able to incorporate the color and relied only on the texture.\\
In our SeaDronesSee experiments, we observe that the gray-scale experiments are outperformed by the color experiments. The explanation can be found in the application.  It covers a maritime environment. As a result, the background is always a mixture of blue and green. So, the color information is reliable in distinguishing between background and foreground. For SeaDronesSee, all models use the color, which is evident in the drop in performance for the gray-scale experiments.\\
If the color is important, RGB seems to produce the overall best results; followed by YCrCb. However, since most data augmentation pipelines have been optimized for RGB, it is recommended to start with RGB.\\
As a conclusion, it is important to know whether color information is relevant in the application. Most times, gray-scale images are sufficient, which leads to a third of the image size and may even allow simpler cameras.

\subsection{Resolution}
For these experiments, we only visualize the experiments of Faster R-CNN. The three selected plots (visible in Fig. \ref{fig:image_size}) are good representatives for the other models, because they describe the general behavior well. The full list of experiments can be found in the supplementary material. \\
First, higher training resolutions always improve the performance of the models. Aside from that, the deterioration of performance for a decreased validation image size is consistent.\\
For the Dota data set, we can observe a performance drop for a higher scale factor. The trained models seem very sensitive to object sizes and cannot take advantage of higher resolutions.\\
For the SeaDronesSee data set, performance improves with higher validation image sizes for most models. So it is possible to improve the performance of a trained model by simply using higher inference resolutions. However, the improvement is marginal and should be supported by scaling data augmentation.\\
The results for the SeaDronesSee data set also apply to the VisDrone data set. Only for the training image size of $1024$ pixels, this rule does not hold. The reason for this is the small number of high-resolution images in the VisDrone validation set. In summary, training and validation size seems to be crucial, and it is not a good idea to cut corners at this point. This also applies for common object, but is even more critical for small object like in this application.

\begin{figure}
    \centering
    \includegraphics[width=0.8\linewidth]{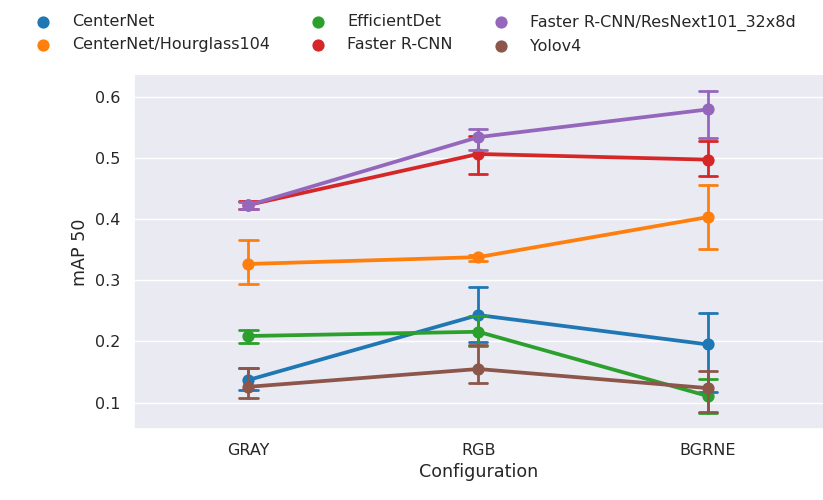}
    \caption{Multispectral: the performance of the object detectors on multispectral data. ('BGRNE' - multispectral recordings, 'RGB' - color images, 'GRAY' - gray-scale)}
    \label{fig:multispectral}
    \vspace{-1em}
\end{figure}

\begin{table*}[t]
\small
\centering

\caption{Application of the optimized parameter configuration in a desktop and an embedded environment.}
\label{tab:application_results}

\setlength\doublerulesep{0.5pt}
\begin{tabular}{llll|lll}
Data set & \multirow{2}{*}{\begin{tabular}[c]{@{}l@{}}Dota \\ mAP 50 $\uparrow$\end{tabular}} & \multirow{2}{*}{\begin{tabular}[c]{@{}l@{}}VisDrone \\ mAP 50 $\uparrow$\end{tabular}} & \multirow{2}{*}{\begin{tabular}[c]{@{}l@{}}SeaDronesSee \\ mAP 50 $\uparrow$\end{tabular}} & \begin{tabular}[c]{@{}l@{}}avg. FPS $\uparrow$ \\ (Desktop)\end{tabular} & \begin{tabular}[c]{@{}l@{}}avg. FPS $\uparrow$ \\ (Embedded)\end{tabular} & \begin{tabular}[c]{@{}l@{}}Parameters\end{tabular} \\
 &  &  &  &  &  &  \\ \hline
\textbf{EfficientDet} & 23.83 & 20.71 & 22.29 & 25 & 14 & \multirow{3}{*}{\begin{tabular}[c]{@{}l@{}}4.5M - \\ 17.7 M\end{tabular}} \\ \cline{1-1}
Reduced data (our) * & 22.07 (-7 \%) & 19.53 (-5 \%) & 22.20 (-1 \%) & \textbf{28 (+12 \%)} & \textbf{16 (+14 \%)} &  \\ \cline{1-1}
\begin{tabular}[c]{@{}l@{}}Reduced data \\ + higher res. (our) **\end{tabular} & \textbf{30.85 (+ 29\%)} & \textbf{25.52 (+23 \%)} & \textbf{27.35 (+23 \%)} & 25 & 14 &  \\ \hline
\textbf{YoloV4} & 16.19 & 24.49 & 40.62 & 20 & 4 & \multirow{2}{*}{63.9M} \\ \cline{1-1}
Reduced data (our) * & 13.84 (-14 \%) & 23.61 (-3 \%) & 36.58  (-9 \%) & \textbf{21 (+5 \%)} & \textbf{5 (+20 \%)} &  \\ \hline
\textbf{CenterNet} & 21.35 & 29.94 & 31.08 & 40 & \multirow{2}{*}{-} & \multirow{2}{*}{\begin{tabular}[c]{@{}l@{}}14.4M - \\ 49.7 M\end{tabular}} \\ \cline{1-1}
Reduced data (our) * & 21.22 (-1 \%) & 27.87 (-6 \%) & 31.05  (-1 \%) & \textbf{46 (+15 \%)} &  &  \\ \hline
\textbf{\begin{tabular}[c]{@{}l@{}}CenterNet/\\ Hourglass104\end{tabular}} & 44.63 & 52.07 & 51.12 & 6 & \multirow{2}{*}{-} & \multirow{2}{*}{200M} \\ \cline{1-1}
Reduced data (our) * & 41.15 (-7 \%) & 48.59 (-6 \%) & 47.89 (-6 \%) & 6 &  &  \\ \hline
\textbf{Faster R-CNN} & 46.67 & 35.45 & 39.94 & 17 & \multirow{2}{*}{-} & \multirow{2}{*}{\begin{tabular}[c]{@{}l@{}}41.4M - \\ 60.3M\end{tabular}} \\ \cline{1-1}
Reduced data (our) * & 43.92 (-5 \%) & 33.05 (-6 \%) & 39.26 (-1 \%) & \textbf{18 (+6 \%)} &  &  \\ \hline
\textbf{\begin{tabular}[c]{@{}l@{}}Faster R-CNN/\\ ResNext101\_32x8d\end{tabular}} & 47.78 & 37.70 & 40.76 & 9 & \multirow{2}{*}{-} & \multirow{2}{*}{104M} \\ \cline{1-1}
Reduced data (our) * & 44.06 (-7 \%) & 35.87 (-4 \%) & 40.59 (-1 \%) & 9 &  &  \\ \hline
\end{tabular}
\end{table*}
\normalsize
\subsection{Multispectral Recordings}
These experiments are based on multispectral data of the SeaDronesSee data set. The set of about 500 multispectral images is not very large, so the results can only give an impression. The usefulness of the additional channels depends strongly on the application. For the maritime environment, the additional information is very useful, especially the near-infrared channel. This part of the spectrum is absorbed by water \cite{Curcio:51}, resulting in a foreground mask (visible in Fig. \ref{fig:example_multispectal}).\\
The results in Fig. \ref{fig:multispectral} fit to the findings for the color models (see section \ref{link:color_model}). For the maritime environment, the color is important, indicated by the gap for gray-scale recordings. The multispectral recordings (labeled as ‘BGRNE’) can only improve the performance of the larger models. For the smaller models, it even leads to a decrease in performance. To support the multispectral recordings as input, we had to increase the input-channels of the first layer to five and lost the pretraining of this layer. The rest of the backbone is still pre-trained.\\
In summary, additional channels can be helpful depending on the use case. It is important to keep in mind that larger backbones can make better use of the additional information. Furthermore, it is a trade-off between a fully pre-trained backbone and additional channels.

\subsection{Camera Calibration}
In our experiments, we are able to observe that the effects of gamma correction and image distortion are equal across all datasets and models. Tab. \ref{fig:gamma_correction} shows, that applying (dynamic) gamma correction to the datasets seems to have a negligible/negative effect on detection accuracy. This implies that gamma correction  indeed does not have a notable effect on detection accuracy.\\
Image distortion seems to also have only a very small effect on detection accuracy, as can be seen in Tab. \ref{fig:image_distortion}. However, the models that have been evaluated on images distorted using pincushion distortion achieve slightly higher accuracy. But although the mean performance over all datasets and models is higher for images with pincushion distortion, most of the models perform best when evacuated on no distortion as can be seen in our supplementary where we provide a detailed analysis over all models and datasets.\\
We can conclude that camera calibration is an existential part in many computer vision applications, however the investigated correction methods seem to only marginally affect the performance of an object detector for remote sensing applications.


\begin{table}[t]
\small

\centering
\caption{Gamma Correction: impact on the performance.}
\label{fig:gamma_correction}
\begin{tabular}{l|l|l|l|l}
 & 0.5 & Orig. & 2.5 & Dynamic \\ \hline
\begin{tabular}[c]{@{}l@{}}Relative\\ mAP50 $\uparrow$\end{tabular} & 98.9 \% & \textbf{100 \%} & 98.3 \% & 99.2 \%
\end{tabular}

\end{table}
\normalsize

\begin{table}[t]
\small

\centering
\caption{Image Distortion: impact on the performance.}
\label{fig:image_distortion}
\begin{tabular}{l|l|l|l}
 & Barrel dist. & Orig. & Pincushion dist. \\ \hline
\begin{tabular}[c]{@{}l@{}}Relative\\ mAP50 $\uparrow$\end{tabular} &  99.8\% &  100 \%&  \textbf{100.6 \%}
\end{tabular}

\end{table}
\normalsize


\section{Application}
\label{link:application}
Based on the previous investigations, we determined the optimal parameter configuration for our application. The goal was to find a suitable compromise between the detection performance and the required space for a recording. The recording size correlates directly with the throughput of the system, including the inference time. A smaller recording size therefore means more frames per second.\\
We considered \textbf{4bit quantization} as a good trade-off. Further, \textbf{JPEG compression quality of 90} seems to be a good compromise. For the urban surveillance use-case, color is not important. Thus, for the \textbf{VisDrone data set, we used gray-scale images}. For the \textbf{other data sets, the color recordings} are required. Finally, the \textbf{highest possible resolution} was used.
In Tab. \ref{tab:application_results}, the application of these findings is visible. The first setup (marked *), which uses the reduced quantization, the JPEG compression and the optional color reduction, increases the throughput (including the inference, the pre- and post-processing) of the smaller models significantly. For the larger models, inference accounts for most of the throughput, so there is no significant speed-up.\\
By combining the previous three adjustment and increasing the resolution (marked **), the performance of the detector can be improved while maintaining the base throughput. In these experiments, the usage of CroW \cite{Varga_2021_ICCV} allowed training with higher resolution.\\
Finally, we use the findings to deploy a UAV based prototype (displayed in Fig. \ref{fig:drone_system}) with nearly real-time and high-resolution processing. The system was carried by a DJI Matrice 100 drone. An Nvidia Xavier AGX board, mounted on the drone, performed the calculations and an Allied Vision 1800 U-1236 camera was used for capturing.

\section{Conclusion}
This paper presents an extensive study of parameters that should be considered when designing a full object detection pipeline in an UAV use-case. Our experiments show that not all parameters have an equal impact on detection performance, and a trade-off can be made between detection accuracy and data throughput.\\
By using the above recommendations, it is possible to maximize the efficiency of the object detection pipeline in a remote sensing application.\\
In a future work, we would like to explore recording data with varying camera parameters in mind. This would give the possibility of evaluating even more camera and image parameters like camera focus or exposure. Also, a deeper analysis of the usability of multispectral imaging could provide useful insights.

\newpage

{\small
\bibliographystyle{ieee_fullname}
\bibliography{egbib}
}

\title{Supplementary material for 'Comprehensive Analysis of the Object Detection Pipeline on UAVs'}

\author{Leon Amadeus Varga\\
Cognitive Systems\\
University of Tuebingen, Germany\\
{\tt\small leon.varga@uni-tuebingen.de}
\and
Sebastian Koch\\
Cognitive Systems\\
University of Tuebingen, Germany\\
{\tt\small se.koch@student.uni-tuebingen.de}
\and
Andreas Zell\\
Cognitive Systems\\
University of Tuebingen, Germany\\
{\tt\small andreas.zell@uni-tuebingen.de}
}

\maketitle



The supplementary material contains additional information and example images for the camera parameters. Further, the experiment setups are described in more detail. We also point to the full list of experiments and code for a preprocessing pipeline, which allows evaluating the parameters in other applications. Both can be found on \ifwacvfinal \url{https://github.com/cogsys-tuebingen/cp_eval} \else [Removed for blind review] \fi.

\section{Experiments}
In this section, we describe the setup of the experiments. The setup contains the used datasets, the object detector models and the hardware environment. For each parameter configuration, we trained with three different random seeds. The random seed affects the weight initialization, the order of the training samples and data augmentation techniques. The three initialization show the stability of parameter configurations.

\subsection{Datasets}
We evaluated our experiments on three data sets. All data sets are part of the remote sensing scenario. DOTA-2 is based on satellites recordings \cite{Ding2021}. VisDrone \cite{DBLP:journals/corr/abs-2001-06303} and SeaDronesSee \cite{Varga2021a} were recorded by Micro Air Vehicles (MAVs). The first two datasets are well established and commonly used to benchmark object detection for remote sensing. In contrast, SeaDronesSee is not completely established, but it focuses on a maritime environment, which is very different to the urban environment of VisDrone and DOTA-2. Therefore, it leads to other challenges for the object detectors and is a proper extension for our experiments.

\paragraph{DOTA-2}
Ding \etal provide high-resolution satellite images with annotations \cite{Ding2021}. We used the recommended preprocessing technique to create 1024 $\times$ 1024 tiles of the high-resolution recordings. All the recordings are provided in the lossless PNG format. Some images contain JPEG compression artifacts, so we assume that not the entire dataset creation pipeline was lossless.

\paragraph{SeaDronesSee-DET}
SeaDronesSee focus on search and rescue in maritime environments \cite{Varga2021a}. The training set includes 2,975 images with 21,272 annotations. We reduced the maximum side length of the images to 1024 pixels for this data set.\\
The reported accuracy is evaluated on the test set (with 1,796 images). This data set uses the PNG format \cite{DBLP:journals/rfc/rfc2083}, which utilizes a lossless compression.

\paragraph{VisDrone-DET}
Zhu \etal proposed one of the most prominent UAV recordings data set \cite{DBLP:journals/corr/abs-2001-06303}. The focus of this data set is traffic surveillance in urban areas. We use the detection task (VisDrone-DET). The training set contains 6,471 images with 343,205 annotations. Unless otherwise mentioned, we reduced the maximum side length by down-scaling to 1024 pixels.\\
We used the validation set for evaluation. The validation set contains 548 images. All images of this data set are provided in JPEG format \cite{DBLP:journals/cacm/Wallace91} with a compression quality of 95.

\subsection{Models}
For our experiments, we selected four different object detectors. The two-stage detector Faster-RCNN \cite{DBLP:journals/pami/RenHG017} can still achieve state-of-the-art results with smaller modifications \cite{DBLP:journals/corr/abs-2001-06303}. By way of comparison, the one-stage detectors are often faster than the two-stage detectors. Yolov4 \cite{Bochkovskiy2020} and EfficentDet \cite{DBLP:conf/cvpr/TanPL20} are one-stage detectors and focus on efficiency. Both are designed to perform well on embedded hardware. Modified versions of CenterNet \cite{Duan2019}, also a one-stage detector, could achieve satisfying results in the VisDrone challenge \cite{DBLP:journals/corr/abs-2001-06303}. These four models cover a variety of approaches and can give a reliable impression of the parameter impact.\\
For Faster R-CNN, EfficentDet and CenterNet, we used multiple backbones to provide an insight for different network sizes. In the following paragraphs, the networks and training setups are described in further detail.\\

\begin{figure*}
    \centering
    \includegraphics[width=0.925\textwidth]{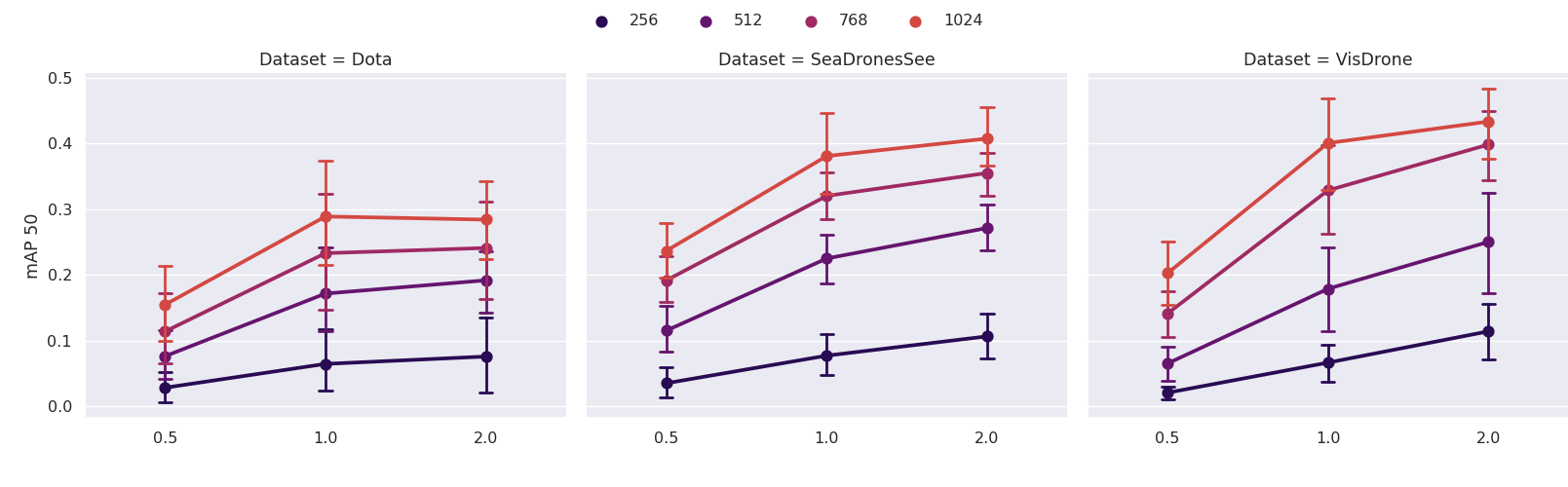}
    \caption{Image Size: CenterNet trained with different training image sizes and validated on $\text{validation image size} = x \times \text{training image size} $.  The y-axis shows the mean Average Precision (mAP).}
    \label{fig:image_size_centernet}
\end{figure*} 
\begin{figure*}
    \centering
    \includegraphics[width=0.925\textwidth]{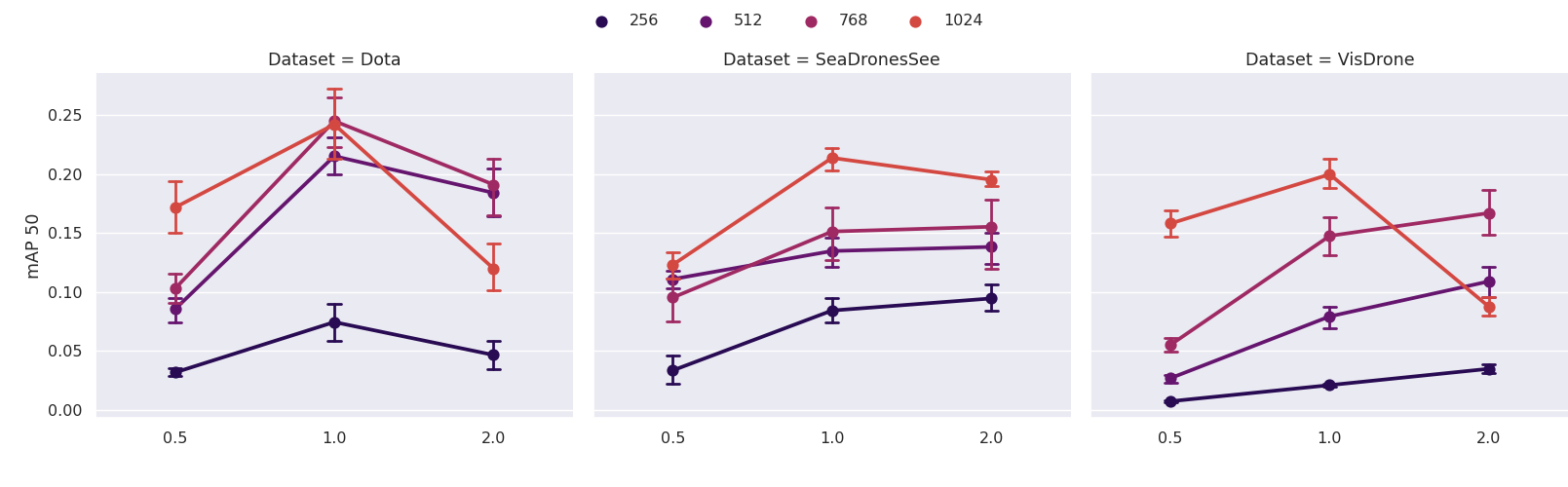}
    \caption{Image Size: EfficientDet trained with different training image sizes and validated on $\text{validation image size} = x \times \text{training image size} $.  The y-axis shows the mean Average Precision (mAP).}
    \label{fig:image_size_efficientdet}
\end{figure*}
\begin{figure*}
    \centering
    \includegraphics[width=0.925\textwidth]{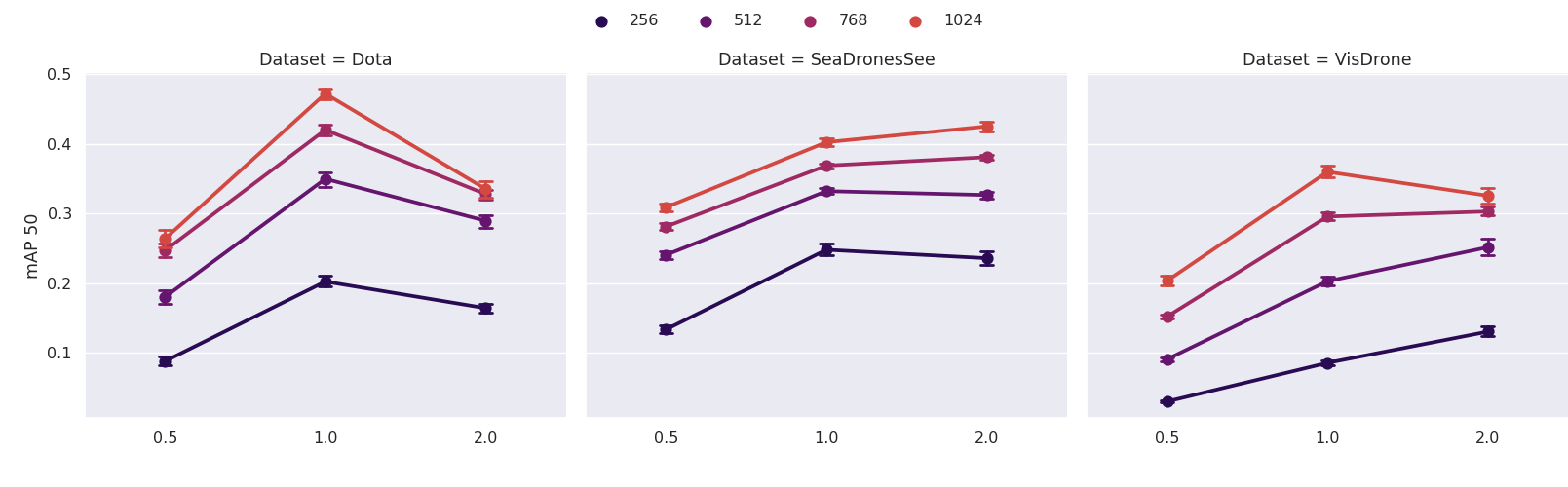}
    \caption{Image Size: Faster R-CNN trained with different training image sizes and validated on $\text{validation image size} = x \times \text{training image size} $.  The y-axis shows the mean Average Precision (mAP).}
    \label{fig:image_size_fasterrcnn}
\end{figure*}
\begin{figure*}
    \centering
    \includegraphics[width=0.925\textwidth]{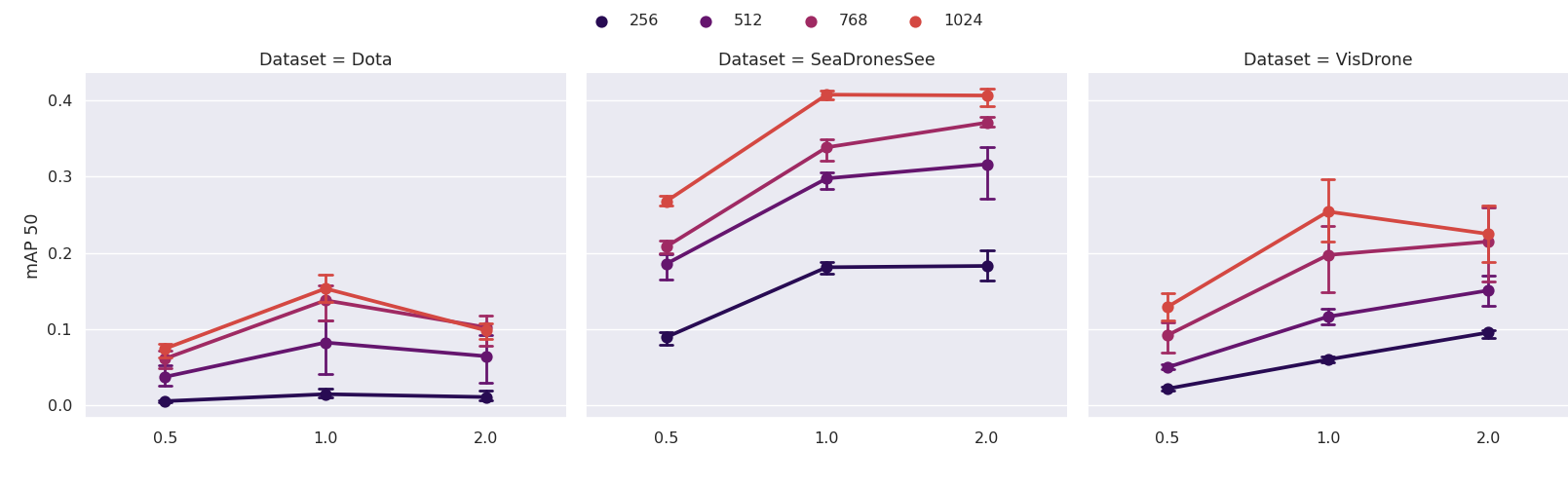}
    \caption{Image Size: YoloV4 trained with different training image sizes and validated on $\text{validation image size} = x \times \text{training image size} $.  The y-axis shows the mean Average Precision (mAP).}
    \label{fig:image_size_yolo}
\end{figure*}

\paragraph{CenterNet}
Duan \etal proposed CenterNet \cite{Duan2019}, an anchor-free object detector. The network uses a heat-map to predict the center-points of the objects. Based on these center-points, the bounding boxes are regressed.
Hourglass-104 \cite{Newell2016StackedEstimation} is a representative for extensive backbones, while the ResNet-backbones \cite{He2016DeepRecognition} cover a variety of different backbone sizes.
The ResNet backbones were trained with Adam and a learning rate of $1e^{-4}$. Further, we also used the plateau learning scheduler. For the Hourglass104-backbone, we used the learning schedule proposed by Pailla \etal \cite{Pailla2019}.

\paragraph{EfficientDet}
EfficientDet is optimized for efficiency and can perform well with small backbones \cite{DBLP:conf/cvpr/TanPL20}. Even there is an EfficentNetV2 announced, which should be more efficient as backbone \cite{DBLP:conf/icml/TanL21}, there is currently, to the best of our knowledge, no object detector using this backbone.
For our experiments, we used EfficientDet with two backbones. The $d0$-backbone is the smallest and fastest of this family. And the $d4$-backbone represents a good compromise between size and performance.
We used three anchor scales (0.6, 0.9, 1.2), which are optimized to detect the small objects of the data sets. For the optimization, we used an Adam optimizer \cite{Kingma2015} with a learning rate of $1e^{-4}$. Further, we used a learning rate scheduler, which reduces the learning rate on plateaus with patience of 3 epochs.

\paragraph{Faster R-CNN}
Faster R-CNN is the most famous two-stage object detector \cite{DBLP:journals/pami/RenHG017}. And many of its improvements achieve today still state-of-the-art results \cite{DBLP:journals/corr/abs-2001-06303}. We use three backbones for Faster R-CNN. A ResNet50 and a ResNet101 of the ResNet-family \cite{He2016DeepRecognition}. Also, a ResNeXt101 backbone \cite{xie2017aggregated}
For Faster R-CNN, we use the Adam optimizer \cite{Kingma2015} with a learning rate of $1e^{-4}$ and a plateau scheduler.

\paragraph{Yolov4}
Bochkovskiy \etal published YoloV4 \cite{Bochkovskiy2020}, which is the latest member of the Yolo-family providing a scientific publication. Besides a comprehensive architecture and parameter search, they did an in-depth analysis of augmentation techniques, called 'bag of freebies', and introduced the Mosaic data augmentation technique.
YoloV4 is a prominent representative of the object detectors because of impressive results on MS COCO. By default, YoloV4 scales all input images down to an image size of 608$\times$608 pixels. For our experiments, we removed this preprocessing to improve the prediction of smaller objects.

\subsection{Hardware setup}
In the following, we describe the setups, which were used for the experiments. Most of the experiments were done in a desktop environment. To evaluate the performance in a more applicable environment, we performed experiments on an Nvidia Xavier AGX board. The small size and the low weight allows the usage for onboard processing in MAVs or other robots.

\paragraph{Desktop environment}
The models were trained on computing nodes equipped with four GeForce GTX 1080 Ti graphic cards. The system was based on the Nvidia driver (version 460.67), CUDA (version 11.2) and PyTorch (version 1.9.0). To evaluate the inference speed in a desktop environment, we used a single RTX 2080 Ti with the same driver configuration. 

\paragraph{Embedded environment}
The Nvidia Xavier AGX provides 512 Volta GPU cores, which are the way to go for the excessive forward passes of the neural networks. Therefore, it is a flexible way to bring deep learning approaches into robotic systems. We used the Nvidia Jetpack SDK in the version 4.6, which is shipped with TensorRT 8.0.1. TensorRT can speed up the inference of trained models. It optimizes these for the specific system, making it a helpful tool for embedded environments. For all embedded experiments, the Xavier board was set to the power mode 'MAXN', which consumes around 30W and utilizes all eight CPUs. Further, it makes use of the maximum GPU clock rate of around 1377 MHz.

\section{Example images for the considered factors of influence}
Further example images for the considered factors of influence can be found here.
\subsection{Quantization}
\begin{figure}
 \begin{subfigure}{0.30 \columnwidth}
        \centering
        \includegraphics[height=3cm]{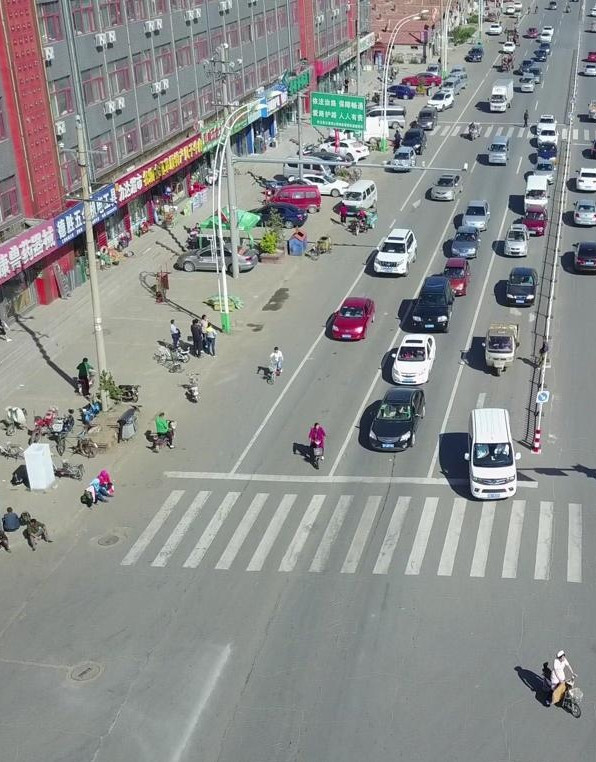}
        \caption{8 Bit}
    \end{subfigure}
    \begin{subfigure}{0.30 \columnwidth}
        \centering
        \includegraphics[height=3cm]{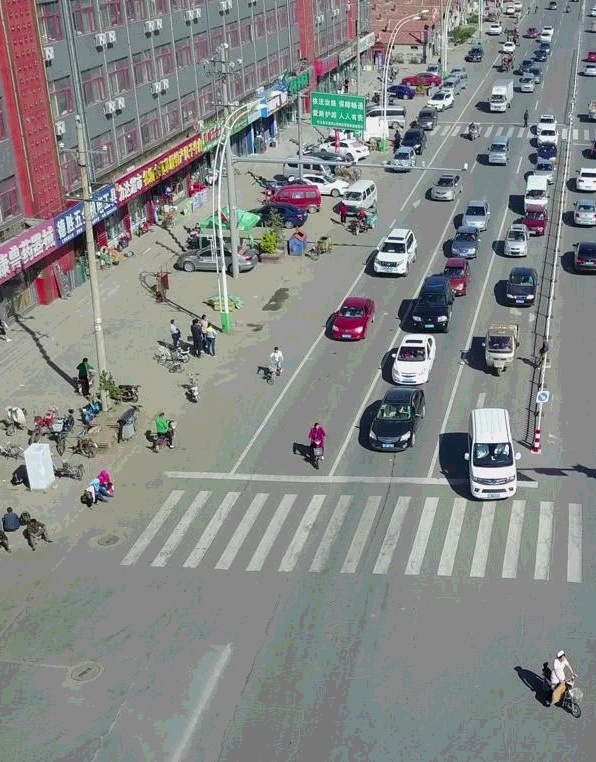}
        \caption{4 Bit}
    \end{subfigure}
    \begin{subfigure}{0.30 \columnwidth}
        \centering
        \includegraphics[height=3cm]{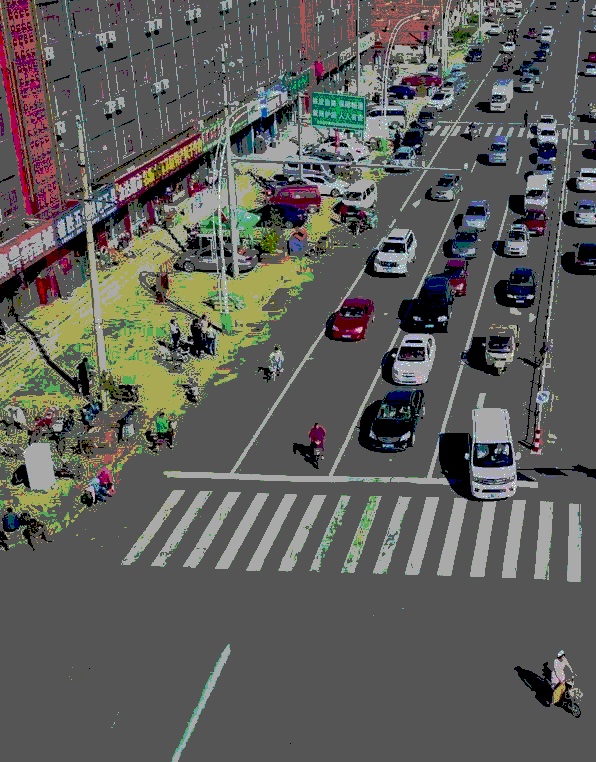}
        \caption{2 Bit}
    \end{subfigure}
    \centering
    \caption{Example images for Quantization}
    \label{fig:example_quantization_visdrone}
\end{figure}

In Fig. \ref{fig:example_quantization_visdrone}, the missing quantization steps for 4 Bit and 2 Bit are visible by the human eye. Already, the 4bit quantization seems not smooth with unwanted steps.
\subsection{Compression}
\begin{figure}
 \begin{subfigure}{0.30 \columnwidth}
        \centering
        \includegraphics[height=3cm]{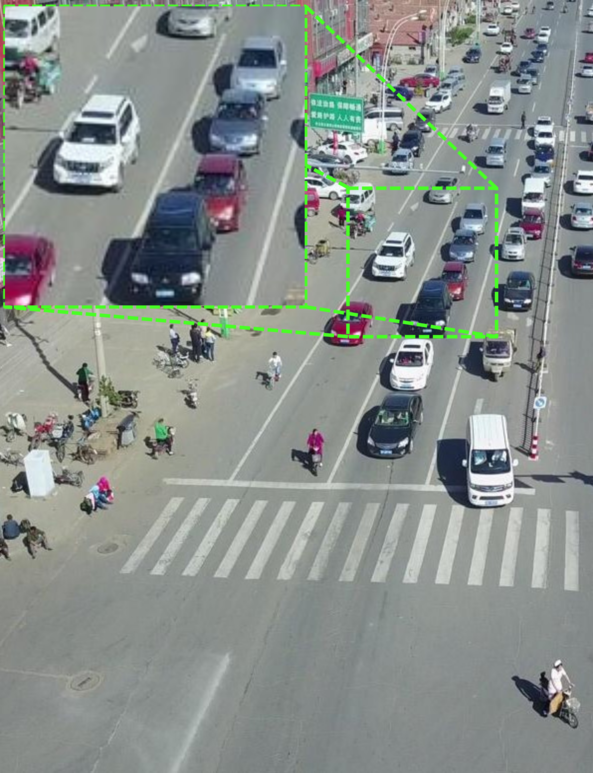}
        \caption{No comp.}
    \end{subfigure}
    \begin{subfigure}{0.30 \columnwidth}
        \centering
        \includegraphics[height=3cm]{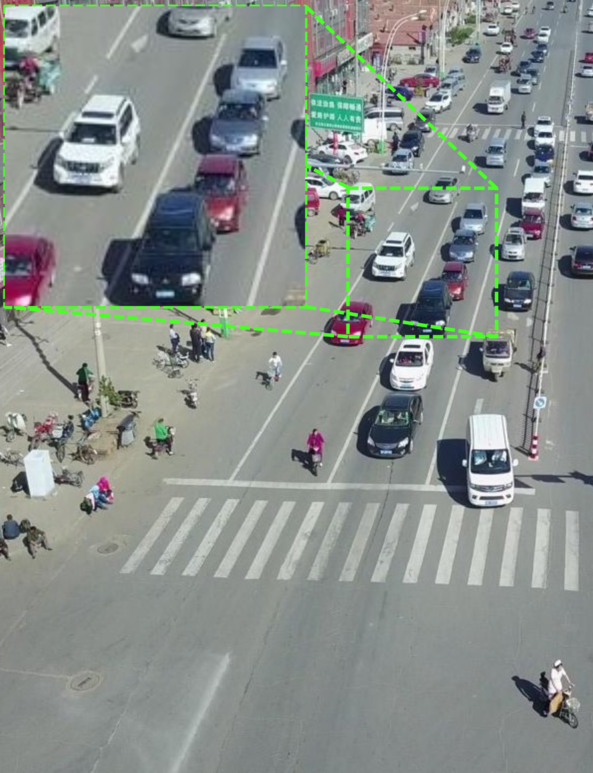}
        \caption{90 comp. qlty.}
    \end{subfigure}
    \begin{subfigure}{0.30 \columnwidth}
        \centering
        \includegraphics[height=3cm]{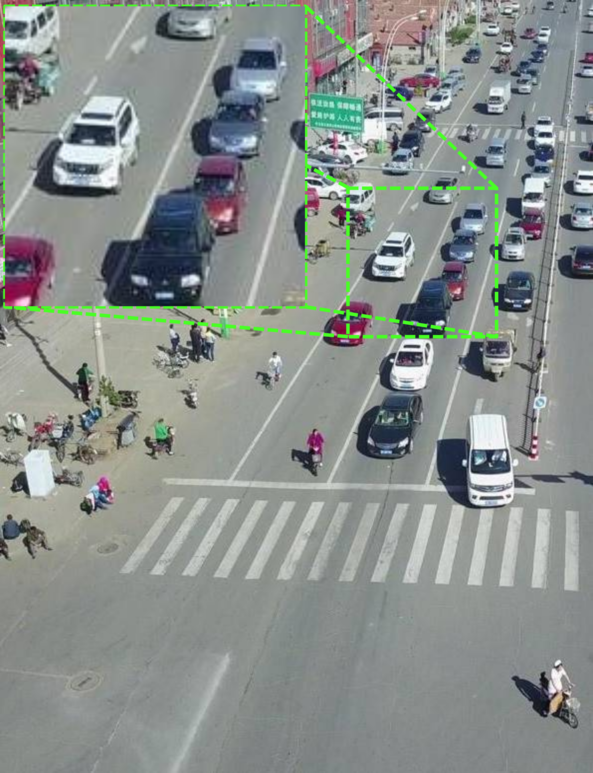}
        \caption{70 comp. qlty.}
    \end{subfigure}
    \begin{subfigure}{0.30 \columnwidth}
        \centering
        \includegraphics[height=3cm]{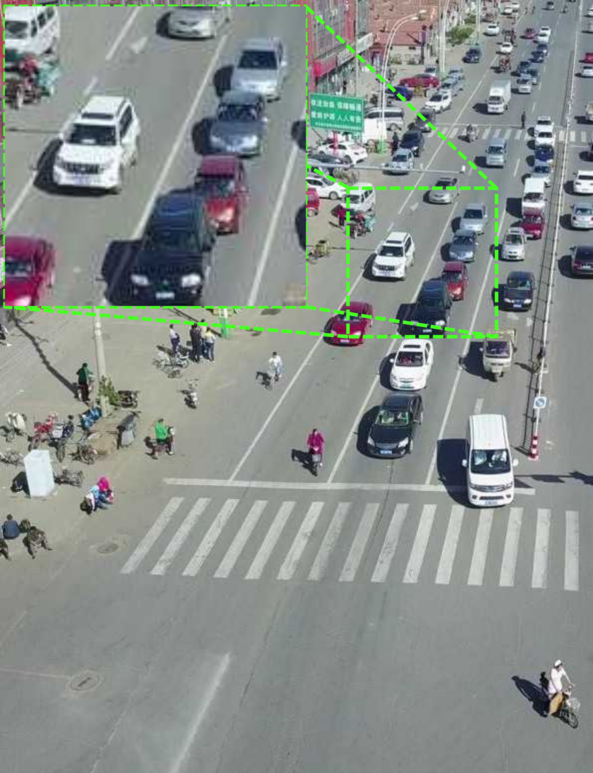}
        \caption{50 comp. qlty.}
    \end{subfigure}
    \begin{subfigure}{0.30 \columnwidth}
        \centering
        \includegraphics[height=3cm]{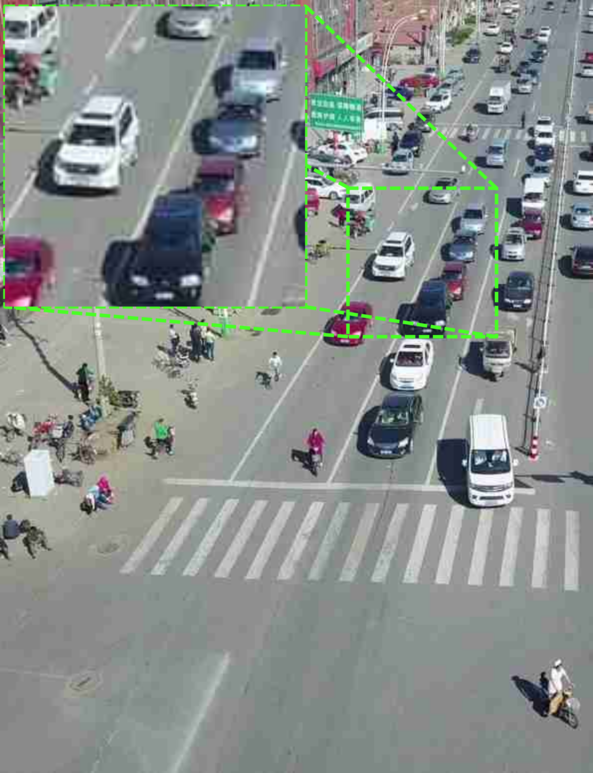}
        \caption{20 comp. qlty.}
    \end{subfigure}
    \begin{subfigure}{0.30 \columnwidth}
        \centering
        \includegraphics[height=3cm]{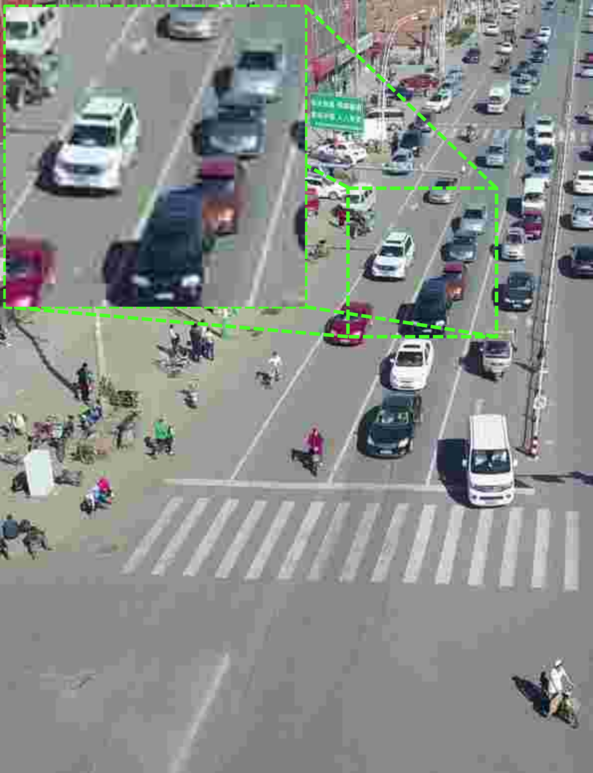}
        \caption{10 comp. qlty.}
    \end{subfigure}
    \centering
    \caption{Example images for JPEG compression}
    \label{fig:example_compression_visdrone}
\end{figure}

In Fig. \ref{fig:example_compression_visdrone} the influence of different compression qualities is visible. For the compression qualities, which were used for the evaluation (\textit{No compression}, \textit{90}, \textit{70}), it is hard to see differences with the human eye. Therefore, the additional compression qualities (\textit{50}, \textit{20}, \textit{10}) are presented. Especially, for the lowest used compression quality (\textit{10}) the typical JPEG compression artifacts are visible (unsharp corners, squares with incorrect color).
                                                                                              
\subsection{Calibration Parameters}     
The amount of distortion is visualized in the figure using a checkerboard pattern.
\begin{figure}[h]
    \centering
    \includegraphics[width=0.9\linewidth]{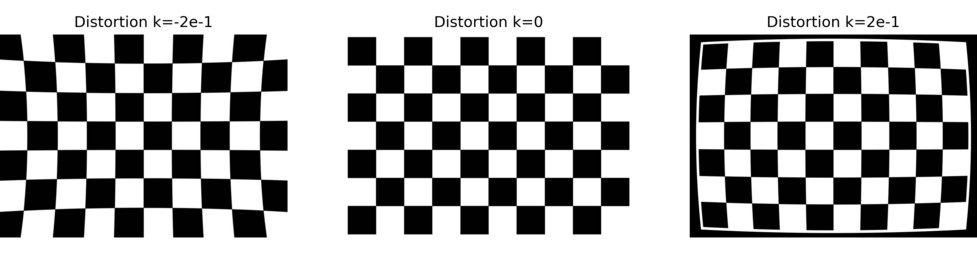}
    \caption{Checkboard pattern with distortion}
    \label{fig:distorion_viz}
\end{figure}

\begin{figure}[h]
 \begin{subfigure}{0.30 \columnwidth}
        \centering
        \includegraphics[height=3cm]{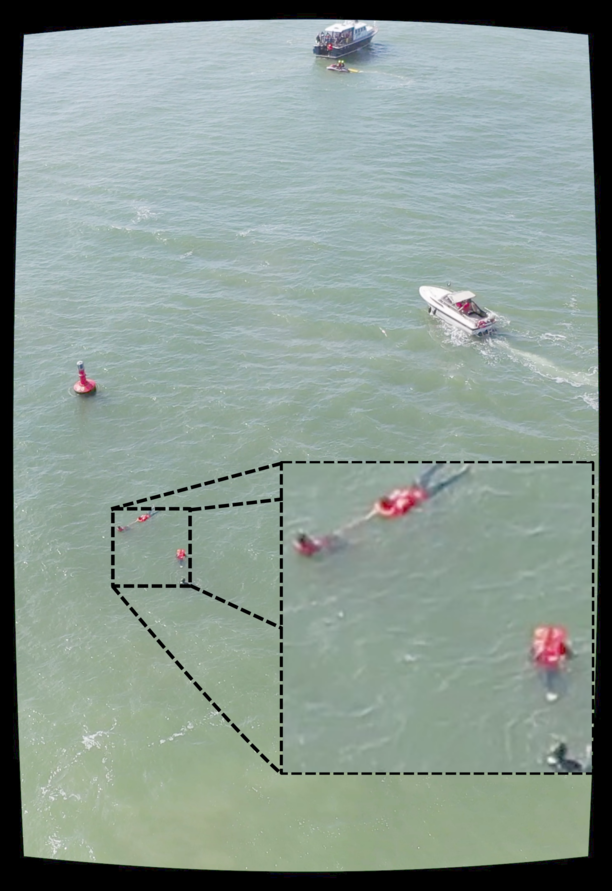}
        \caption{}
    \end{subfigure}
    \begin{subfigure}{0.30 \columnwidth}
        \centering
        \includegraphics[height=3cm]{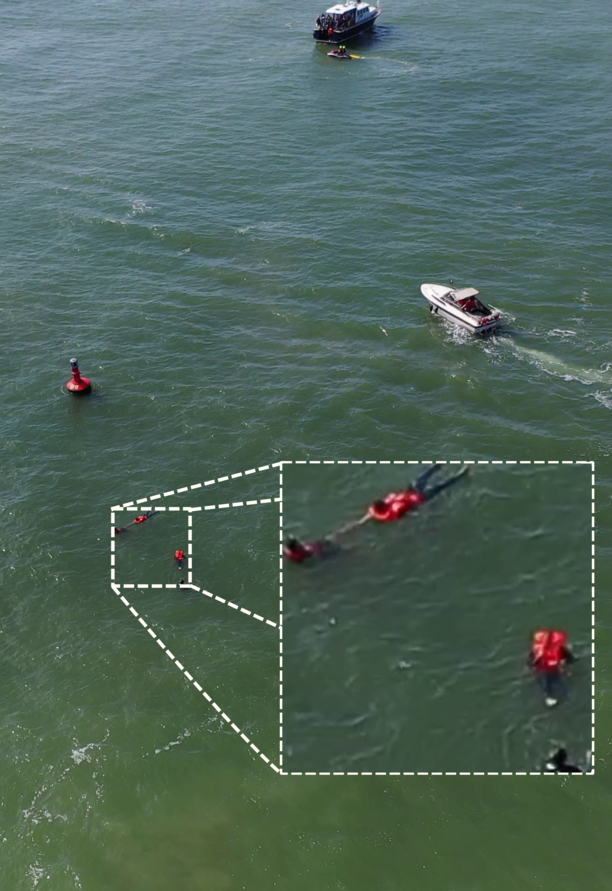}
        \caption{Rectified}
    \end{subfigure}
    \begin{subfigure}{0.30 \columnwidth}
        \centering
        \includegraphics[height=3cm]{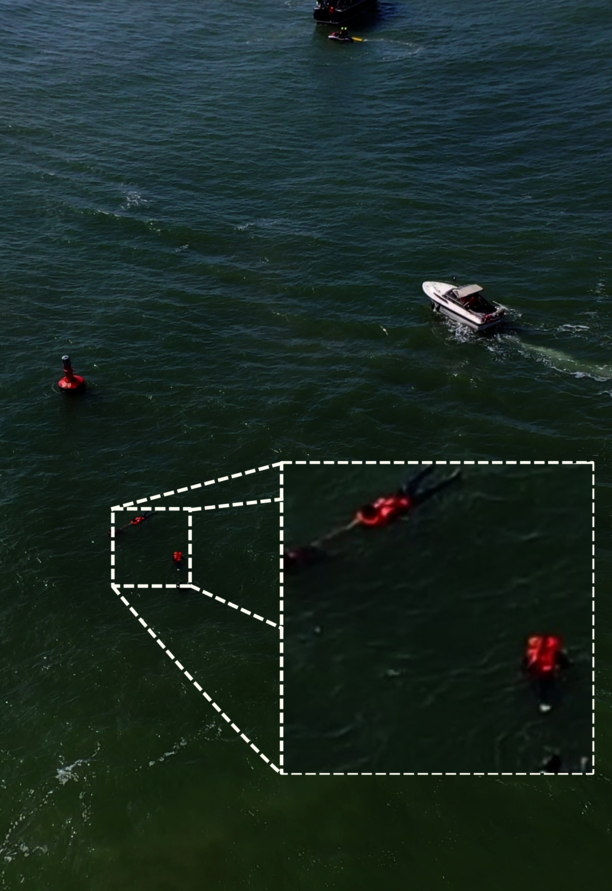}
        \caption{}
    \end{subfigure}
    \centering
    \caption{Example images for Camera Calibration}
    \label{fig:example_calibration_parameter}
\end{figure}

\begin{figure*}
    \centering
    \includegraphics[width=0.925\textwidth]{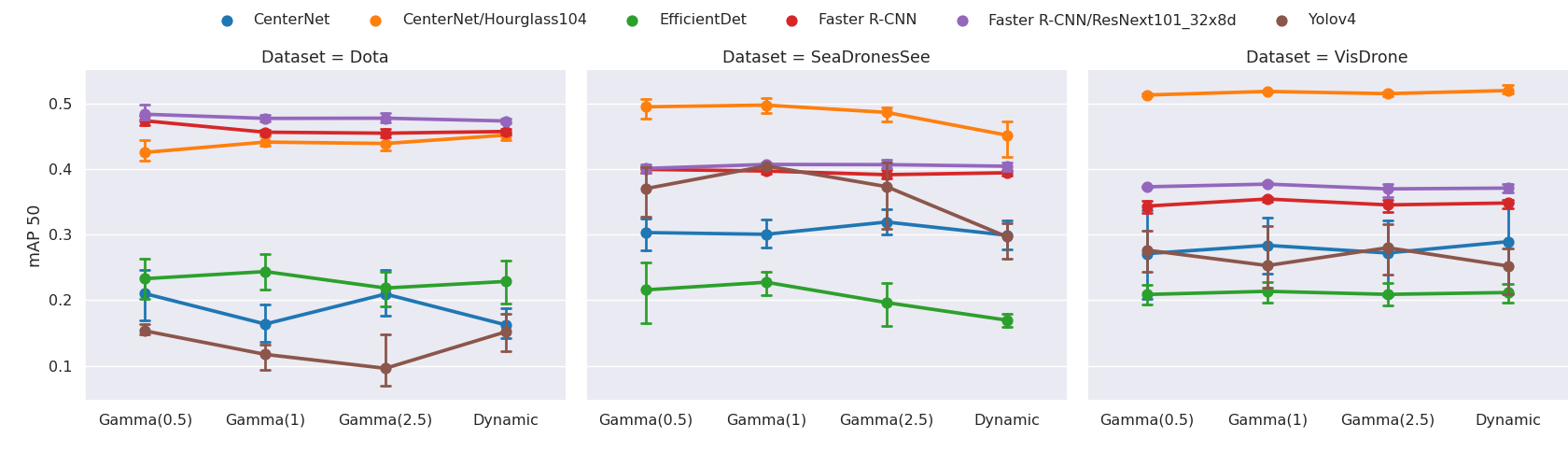}
    \caption{Image Gamma: the impact of the gamma on the performance of the object detectors for all data sets. The y-axis shows the mean Average Precision (mAP).}
    \label{fig:image_gamma}
\end{figure*}

The Fig. \ref{fig:example_calibration_parameter} shows the rectified image and two not optimal setups. In (a), the image is overexposed and the barrel distortion of the lens was not fully corrected. In setup (c), the image is underexposed, and the recording is affected by a pincushion distortion. In a real world image, the distortions are hard to see. They are only visible for straight lines. In contrast, the incorrect exposure, which correlates with the gamma value, is obvious. In Fig. \ref{fig:image_distortion} the breakdown of the image distortion impact on the different data sets and models is visualized. Fig. \ref{fig:image_gamma} shows the same for the impact of gamma.

\begin{figure*}
    \centering
    \includegraphics[width=0.925\textwidth]{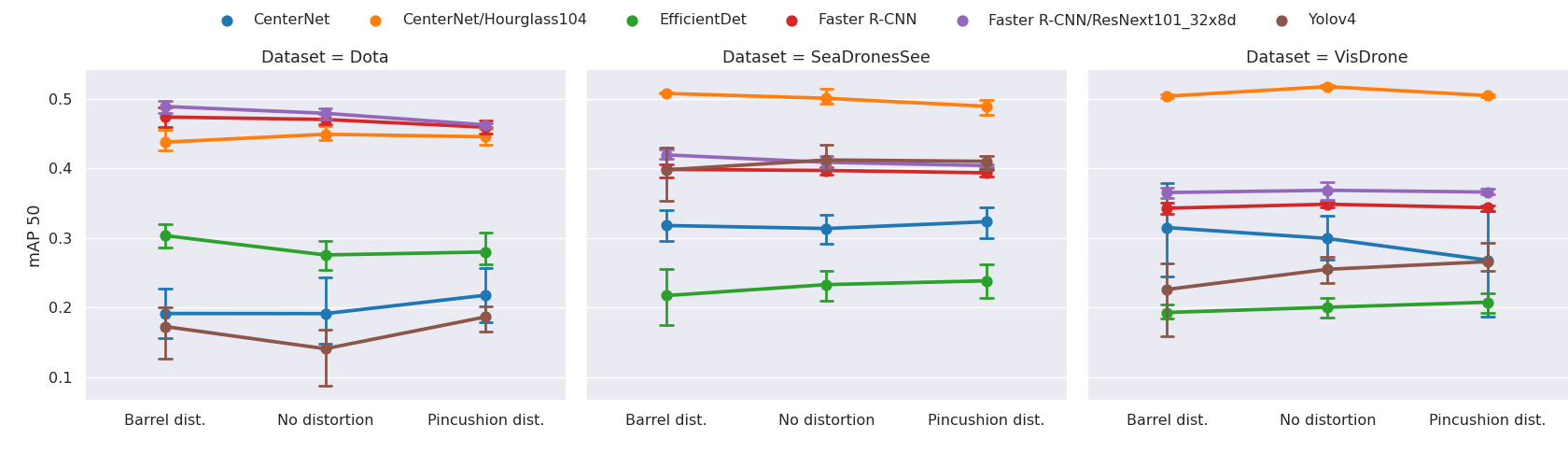}
    \caption{Image Distortion: the impact of the image distortion on the performance of the object detectors for all data sets. The y-axis shows the mean Average Precision (mAP).}
    \label{fig:image_distortion}
\end{figure*}

\begin{figure}
 \begin{subfigure}{0.30 \columnwidth}
        \centering
        \includegraphics[height=3cm]{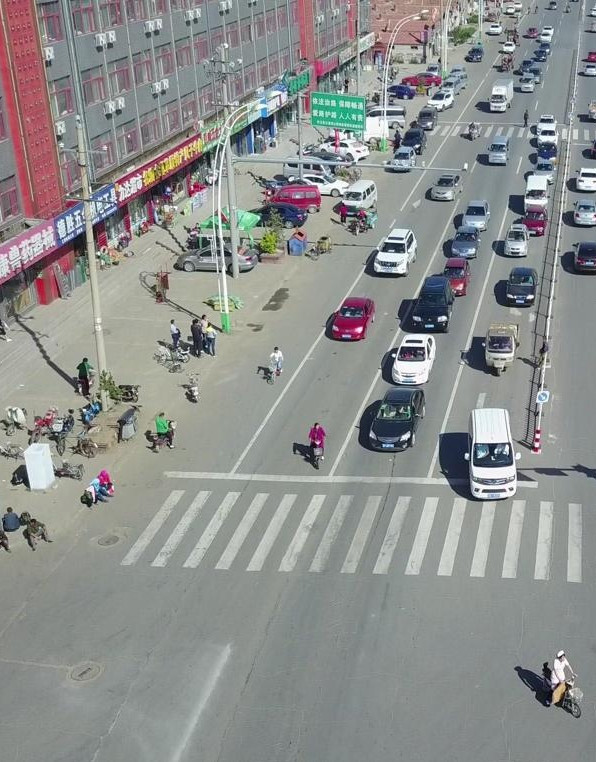}
        \caption{Full res.}
    \end{subfigure}
    \begin{subfigure}{0.30 \columnwidth}
        \centering
        \includegraphics[height=3cm]{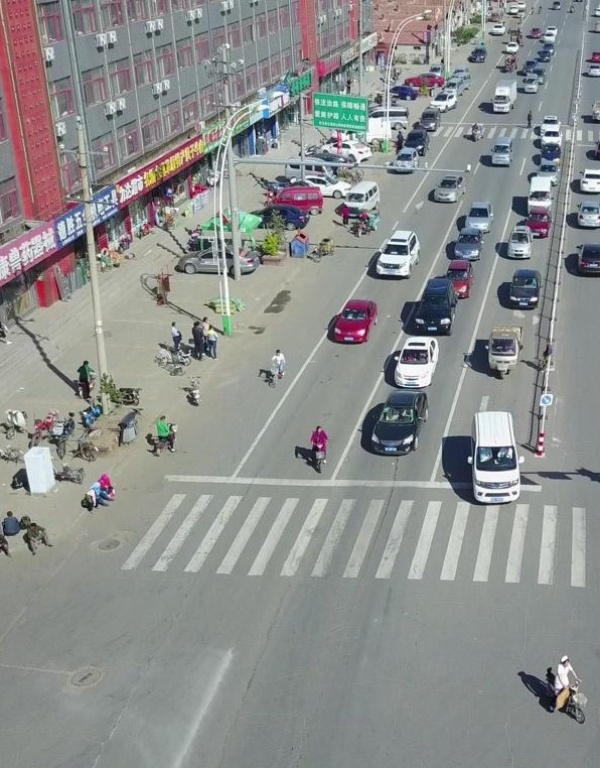}
        \caption{Max 768 pix.}
    \end{subfigure}
    \begin{subfigure}{0.30 \columnwidth}
        \centering
        \includegraphics[height=3cm]{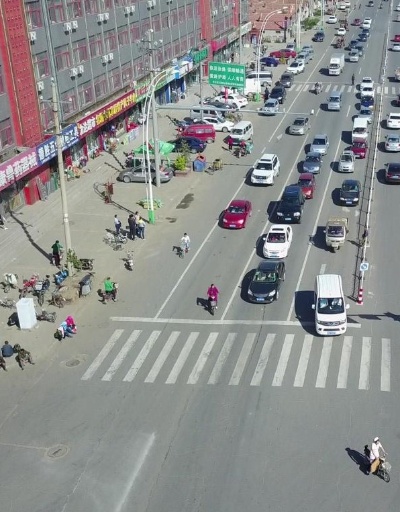}
        \caption{Max 512 pix.}
    \end{subfigure}
    \begin{subfigure}{0.30 \columnwidth}
        \centering
        \includegraphics[height=3cm]{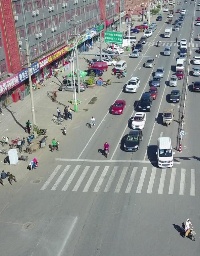}
        \caption{Max 256 pix.}
    \end{subfigure}
    \begin{subfigure}{0.30 \columnwidth}
        \centering
        \includegraphics[height=3cm]{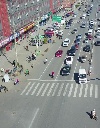}
        \caption{Max 128 pix.}
    \end{subfigure}
    \centering
    \caption{Example images for image resolutions with max side length}
    \label{fig:example_resolution_visdrone}
\end{figure}

\subsection{Resolution}
In Fig. \ref{fig:example_resolution_visdrone} an image with different resolutions is shown. For smaller resolution, the detection of smaller objects is especially hard. These can vanish completely. Large and medium objects normally lose only features, which makes the detection harder but not impossible. Therefore, the impact on smaller objects is higher.

\section{Results and discussion}
The full list of experiments can be found on \ifwacvfinal \url{https://github.com/cogsys-tuebingen/cp_eval} \else [Removed for blind review] \fi.
Here are some additional information regarding the experiment results.
                       
\subsection{Quantisation}
%
%

\subsection{Resolution}
In Fig. \ref{fig:image_size_centernet}, Fig. \ref{fig:image_size_efficientdet}, Fig. \ref{fig:image_size_fasterrcnn} and Fig. \ref{fig:image_size_yolo}, the behavior of the different models for the tested image resolutions is visible. All models follow the trend of Faster RCNN, which was described in the paper. EfficientDet seems more sensitive to resolution changes than the other models.

\subsection{Gamma Correction}
\subsection{Image Distortion}

\section{Application}
For the **) evaluations, we used a maximal side length of 1280 pixels. We used therefore the CroW technique \cite{Varga_2021_ICCV} in the default configuration. These are the only changes for this evaluation.

{\small
\bibliographystyle{ieee_fullname}
\bibliography{egbib}
}

\end{document}